\documentclass{bmvc2k}

\usepackage[utf8]{inputenc}
\usepackage{xcolor}
\usepackage{pifont}
\usepackage{color, colortbl}
\usepackage{algorithm}
\usepackage{algorithmic}
\usepackage{multirow}
\usepackage{times}
\usepackage{epsfig}
\usepackage{lipsum} 
\usepackage{url}

\newcommand{\cmark}{\ding{51}}%
\newcommand{\xmark}{\ding{55}}%

\title{What Should Be Balanced in a "Balanced" Face Recognition Dataset?}

\addauthor{Haiyu Wu}{hwu6@nd.edu}{1}
\addauthor{Kevin W. Bowyer}{kwb@nd.edu}{1}

\addinstitution{
 University of Notre Dame\\
 Notre Dame, USA
}

\runninghead{Wu et al}{What Should be Balanced in a "Balanced" Dataset?}

\hypersetup{
urlcolor=blue,
}
\begin{document}

\maketitle

\begin{abstract}
The issue of demographic disparities in face recognition accuracy has attracted increasing attention in recent years. Various face image datasets have been proposed as 'fair' or 'balanced' to assess the accuracy of face recognition algorithms across demographics. These datasets typically balance the number of identities and images across demographics. It is important to note that the number of identities and images in an evaluation dataset are {\em not} driving factors for 1-to-1 face matching accuracy. Moreover, balancing the number of identities and images does not ensure balance in other factors known to impact accuracy, such as head pose, brightness, and image quality. We demonstrate these issues using several recently proposed datasets. To improve the ability to perform less biased evaluations, we propose a bias-aware toolkit that facilitates creation of cross-demographic evaluation datasets balanced on factors mentioned in this paper. The dataset is at \url{https://github.com/HaiyuWu/BA-test-dataset}.
\end{abstract}

\section{Introduction}
\label{sec:intro}
Demographics disparity in face recognition accuracy  has emerged as a significant and contentious issue~\cite{Lohr2018, NIST_2019, Hoggins2019, drozdowski2020demographic, NIST_2022, krishnenduanalysis}. Researchers have explored approaches to uncover the causes of observed accuracy differences~\cite{wu2023face, bhatta2023gender, 9650887, krishnapriya2020issues, vangara2019characterizing, albiero2020does, albiero2020analysis}.  Datasets have been proposed as `fair' or `balanced' for evaluating accuracy across  groups~\cite{bfw, demogpairs, Wang_2019_ICCV}. In this paper, we argue that merely balancing the number of identities and images is insufficient for establishing a fair evaluation. Instead, we posit that creating a fair evaluation necessitates balancing factors known to impact accuracy, such as image quality~\cite{terhorst2023qmagface}, head pose~\cite{gross2010multi}, brightness~\cite{wu2023face}, and age~\cite{albiero2020does}. We assembled a Bias Aware test set (BA-test) that balances multiple known accuracy-related factors, enabling cross-demographic evaluations with minimal inherent bias. Contributions of this work include:

\begin{itemize}
\vspace{-2mm}
\item We demonstrate that datasets previously deemed "fair" or "balanced" for evaluation across demographics are not balanced on factors known to be essential for accuracy.
\vspace{-2mm}
\item We introduce the BA-test dataset, designed to support demographic accuracy disparity evaluations based on a better-balanced test set.
\vspace{-2mm}
\item We offer a toolkit to balance a given dataset based on all factors balanced in BA-test.
\vspace{-2mm}
\item We provide an accuracy-disparity-focused benchmark, revealing that current state-of-the-art models exhibit  lowest accuracy on Asian females and highest on White males.
\end{itemize}

\section{Literature Review}
\label{sec:related}

AI fairness is currently a topic of significant interest, with datasets~\cite{hazirbas2021towards, porgali2023casual} proposed to measure fairness in audio, vision, and speech domains. Numerous datasets have been introduced to evaluate facial recognition  robustness in various ways. CFP~\cite{cfp} and CPLFW~\cite{cplfw} concentrate on the head pose factor, while CALFW~\cite{calfw} and AgeDB~\cite{agedb} examine accuracy across diverse ages and age gaps. LFW~\cite{lfw}, IJB-C~\cite{ijbc}, and MegaFace~\cite{megaface} pose challenges to models in unconstrained environments with variations in head pose, brightness, expression, age, and image quality. On the other hand, MORPH~\cite{ricanek2006morph}, CMU-PIE~\cite{han2013comparative, Sim2003PAMI}, and Multi-PIE~\cite{gross2010multi} include images captured under more controlled conditions.

To support study of demographic accuracy disparity, Hupont et al.\cite{demogpairs} introduced  DemogPairs, which is balanced with 10.8K images from 600 identities across six demographics. Robbins et al.\cite{bfw} developed Balanced Faces in-the-Wild (BFW), which includes an Indian subgroup and comprises 20K images from 800 identities. Wang et al.\cite{Wang_2019_ICCV} proposed Racial Faces in-the-wild (RFW) with 20K images from 12K subjects across  (White, Black, Asian, Indian). DemogPairs, BFW, and RFW focus on balancing the number of identities or the number of images per identity. BUPT-BalancedFace\cite{wang2021meta} is a training set containing 1.3M images from 28K celebrities across four ethnicities, with 7K identities per ethnicity. However, a fair evaluation of accuracy differences requires more than just balancing the number of subjects and images. Our proposed BA-test dataset balances image quality, head pose, and brightness as examples of controlling factors known to impact recognition accuracy.

Various tools exist to sub-sample datasets to improve performance~\cite{zhu2006subclass, zhu2004optimal}, clean the dataset~\cite{deng2020sub}, correct imbalance~\cite{lemaitre2017imbalanced}, and provide a test set with specified  traits~\cite{narayan2021personalized, wang2021facex, wang2021face, robbins2023cast}. However, none of these tools specifically address factors that directly impact demographic disparity in face recognition accuracy. Therefore, we propose the BA-toolkit to balance selected factors (e.g., brightness, head pose, image quality, age, amount of visible face) within a dataset, enabling a more controlled evaluation of accuracy across demographics.

\vspace{-2mm}
\section{What Matters for a Balanced Test Set?}
\vspace{-2mm}

\label{sec:matters_doesn't_matter}
Existing training~\cite{wang2021meta} and testing~\cite{Wang_2019_ICCV, demogpairs, bfw} datasets balance  number of subjects and number of images per subject. However, number of subjects and images per subject in a test set does not drive differences in 1-to-1 matching accuracy; see Section~\ref{sec:id_im_balanced}.
Factors such as image quality~\cite{terhorst2023qmagface}, head pose~\cite{gross2010multi}, brightness~\cite{wu2023face}, hairstyle~\cite{9650887, bhatta2023gender, wu2023logical}, and facial morphology~\cite{9650887, albiero2020face} can cause accuracy differences across demographics. Moreover, ~\cite{albiero2020does-balance_training, gwilliam2021rethinking} concluded that gender and race balance in  training data does not translate into gender and race balance in  test accuracy. To investigate factors known causes accuracy differences, we use the well-known VGGFace2~\cite{vggface2} dataset, and analyze factors known to impact accuracy.

\vspace{-2mm}
\subsection{Bias Aware Toolkit}
\label{sec:toolkit}
VGGFace2  consists of in-the-wild, uncropped, unaligned images without demographic meta-data and contains noise in identity labels. To support demographic analysis, we propose a Bias Aware toolkit (BA-toolkit) that integrates the function of predicting gender, race, age labels and balancing factors that matter for face recognition accuracy, including brightness, head pose, image quality, and visible face area. The details of the processes are:\\
    \noindent
    \textbf{Data preparation}: Images are cropped and aligned by img2pose ~\cite{albiero2021img2pose} and resized to 224$\times$224. Then the measured 3D head poses are converted from 6DoF to degree in Pitch, Yaw, Roll. To prepare for face brightness measurement and balancing face area across gender,  BiSeNet~\cite{bisenet_github, yu2018bisenet} is used to segment a face image. For image quality, FaceQnet~\cite{faceqnetv1} and MagFace~\cite{magface} are used. FairFace~\cite{karkkainen2019fairface} is used to predict the demographic and age labels of each image, and we make demographic labels consistent for a given identity by voting the race and gender within each identity. To reduce identity noise, we used ArcFace and MagFace to extract features from the selected images and use DBSCAN~\cite{dbscan} to clean label noise.\\
    \noindent
    \textbf{Image selection}: We want to balance  image quality, head pose, and brightness. First, the FaceQnet quality~\cite{faceqnetv1} is used to select images ($Q_{im} >0.3$). To cross-check quality, MagFace is used the drop images with MagFace quality $< 20$. For frontal pose, only  images with $max\{Pitch, Yaw, Roll\}$ $\in [-20^o,20^o]$, are selected. Filtering the image set for the brightness range recommended by~\cite{wu2023face} would reduce the number of images too much. Hence, we use the middle-exposed range $[115.86, 198.75]$ to filter images. To reduce identity noise, we use DBSCAN to get  cosine distance between features in one identity folder and drop outliers.\\

\vspace{-6mm}
\subsection{Preliminary Data Preparation}
\begin{figure}[t]
\centering
    \begin{subfigure}[b]{1\linewidth}
    \centering
        \begin{subfigure}[b]{1\linewidth}
            \begin{subfigure}[b]{0.24\linewidth}
                \begin{subfigure}[b]{0.48\linewidth}
                    \includegraphics[width=\linewidth]{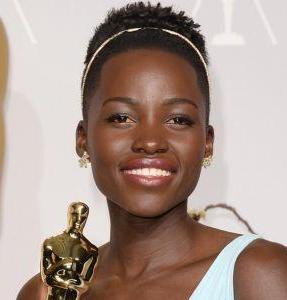}
                \end{subfigure}
                \begin{subfigure}[b]{0.48\linewidth}
                    \includegraphics[width=\linewidth]{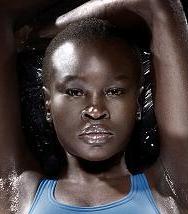}
                \end{subfigure}
            \caption{Black female}
            \end{subfigure}
            \begin{subfigure}[b]{0.24\linewidth}
                \begin{subfigure}[b]{0.48\linewidth}
                    \includegraphics[width=\linewidth]{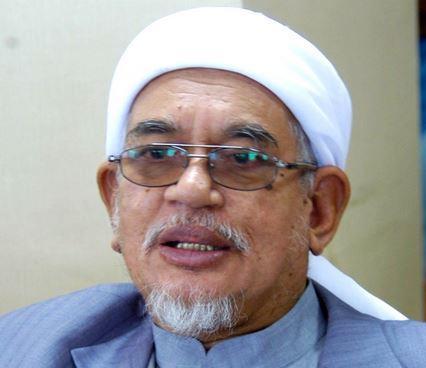}
                \end{subfigure}
                \begin{subfigure}[b]{0.48\linewidth}
                    \includegraphics[width=\linewidth]{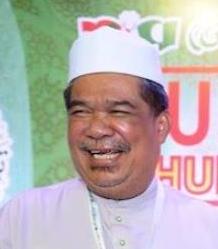}
                \end{subfigure}
            \caption{Asian male}
            \end{subfigure}
            \begin{subfigure}[b]{0.24\linewidth}
                \begin{subfigure}[b]{0.48\linewidth}
                    \includegraphics[width=\linewidth]{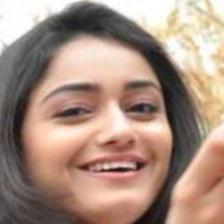}
                \end{subfigure}
                \begin{subfigure}[b]{0.48\linewidth}
                    \includegraphics[width=\linewidth]{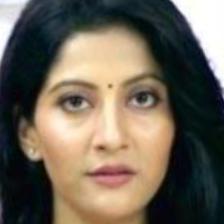}
                \end{subfigure}
            \caption{Indian female}
            \end{subfigure}
            \begin{subfigure}[b]{0.24\linewidth}
                \begin{subfigure}[b]{0.48\linewidth}
                    \includegraphics[width=\linewidth]{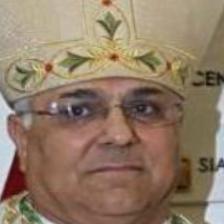}
                \end{subfigure}
                \begin{subfigure}[b]{0.48\linewidth}
                    \includegraphics[width=\linewidth]{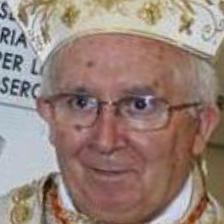}
                \end{subfigure}
            \caption{White male}
            \end{subfigure}
        \end{subfigure}
    \end{subfigure}
   \caption{Identity noise in VGGFace2 -- Each pair is labeled as same identity in  VGGFace2.}
\label{fig:noise_samples}
\vspace{-5mm}
\end{figure}

\begin{figure}[t]
\centering
    \begin{subfigure}[b]{1\linewidth}
        \begin{subfigure}[b]{0.49\linewidth}
            \includegraphics[width=\linewidth]{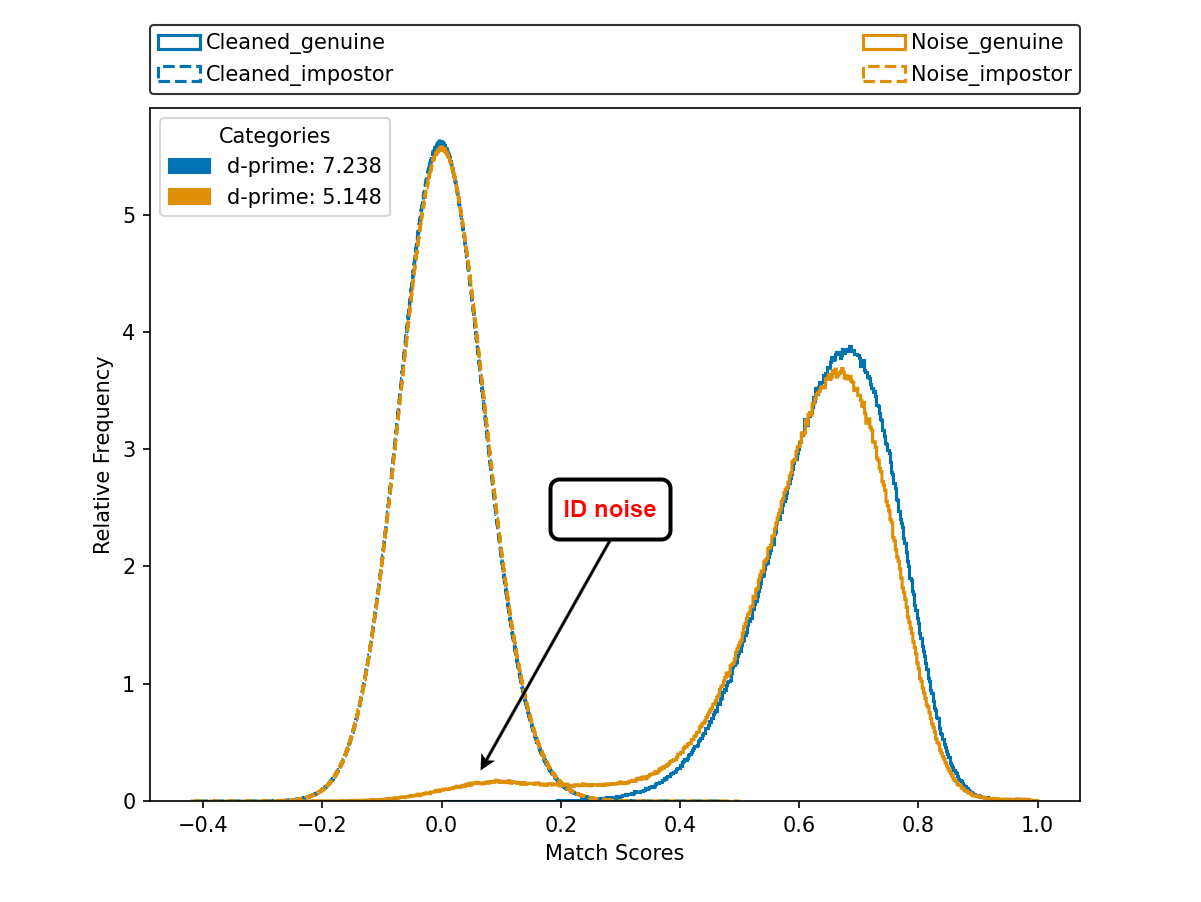}
            \caption{Identity de-noising}
            \label{fig:id_clean}
        \end{subfigure}
        \begin{subfigure}[b]{0.49\linewidth}
            \includegraphics[width=\linewidth]{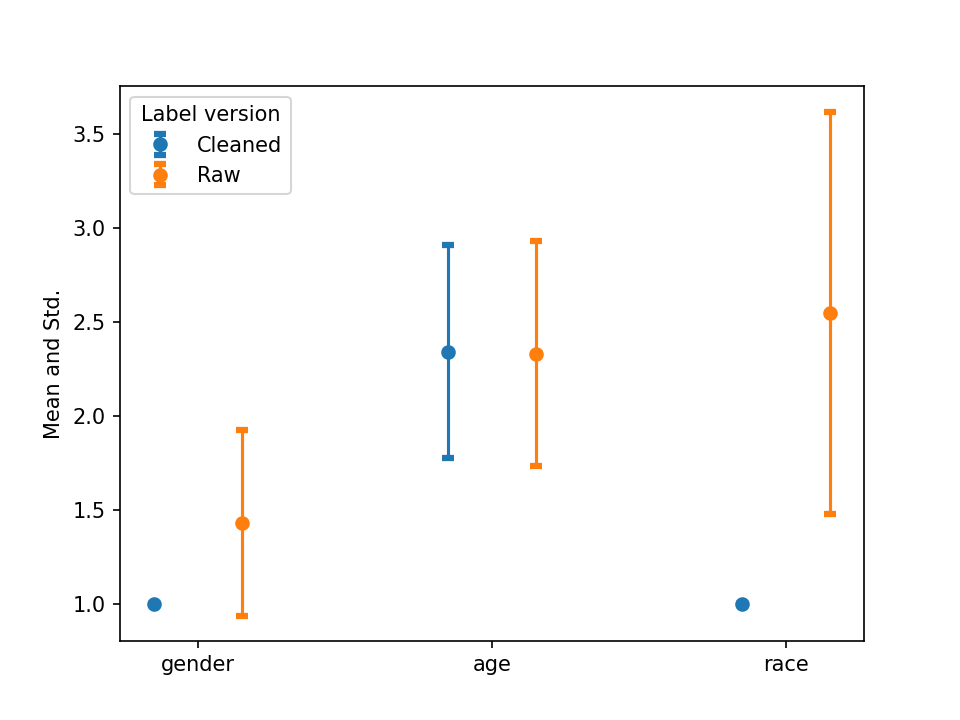}
            \caption{Label cleaning}
            \label{fig:age_gender_race}
        \end{subfigure}
    \end{subfigure}
   \caption{a) Genuine / impostor distributions of random 200 VGGFace2 identities
before and after cleaning identity noise. A  “fair” or “balanced” dataset should
have the same level of clean identity labels across demographics. b) Mean and std. dev. of the number of race, age, gender within each identity before and after cleaning; identity label cleaning results in a single gender and race across the images for a given identity.}
\vspace{-5mm}
\end{figure}

\noindent
\textbf{Quality-driven selection}: We first use VGGFace2 FaceQnet scores~\cite{FQN_github} to drop around 1 million low quality images, a blur classifier to drop 124,632 blurry images, and img2pose to crop and align the the most frontal of the remaining images (dropping 562,036 images).  This leaves 1,286,240 images for next steps.\\
\noindent
\textbf{Intra-class noise}: 
Existing datasets~\cite{Wang_2019_ICCV, bfw, demogpairs} were assembled from resources such as VGGFace2 and MS-Celeb-1M, which have identity noise. Identity noise that varies across demographics can lead to incorrect conclusions about accuracy differences, but previous `balanced' datasets have not cleaned the identity noise. To analyze identity label noise in VGGFace2 we randomly select 200 identities and calculate the cosine similarities. Figure~\ref{fig:id_clean} shows that the genuine distribution before identity cleaning has an impostor-like peak from -0.2 to 0.3. Algorithms exist~\cite{dbscan, deng2020sub} to clean this noise. We implement DBSCAN~\cite{dbscan}, on the features extracted by ArcFace~\cite{arcface, insightface} and MagFace~\cite{magface} in order to minimze the level of identity noise. The genuine distribution of the randomly-selected identities after cleaning (Figure~\ref{fig:noise_samples}) indicates the identity noise is reduced; 49,468 images are dropped in this step.

\noindent
\textbf{Race, Age, gender labels}: The FairFace~\cite{karkkainen2019fairface} classifier, trained on race-balanced data,  provides confidences on four and seven races, nine age intervals, and two genders. To avoid over-classification and ambiguity between race groups (e.g., SE Asian and East Asian), we choose to group the identities in four groups (i.e. White, Black, Indian, Asian). For age, we follow previous works~\cite{albiero2020does, klare2012face-age}, using Young (10-29 since FairFace predicts age chunks), Middle-Aged (30-49) and Senior (50+) groups. Figure~\ref{fig:age_gender_race} shows the variation in the number of race, gender, and age within the identity folders. For the gender and race attributes, if a given identity has more than one predicted value, the values were made consistent using the identity's most frequently occurring value. 
Age can vary across the images of the same identity, so we tried to compare the result with the age classifier $F_{age}$ in~\cite{albiero2020does-balance_training}. These two age predictors disagree on 70K images, where 53K cases are between Young (Fairface) and Middle\_Aged ($F_{age}$). The others are in group (Middle\_Aged, Young), (Middle\_Aged, Senior), (Young, Senior). We manually inspected 2000 random samples with different age predictions from both classifiers and found that FairFace has more accurate age predictions. Consequently, we only use FairFace results and manually adjust the annotations of some evidently incorrect predictions. 146,842 samples underwent label adjustments in this step.

\begin{figure*}[t]
\centering
    \begin{subfigure}[b]{1\linewidth}
    \captionsetup[subfigure]{labelformat=empty}
    \centering
        \begin{subfigure}[b]{1\linewidth}
            \begin{subfigure}[b]{0.24\linewidth}
                \includegraphics[width=\linewidth]{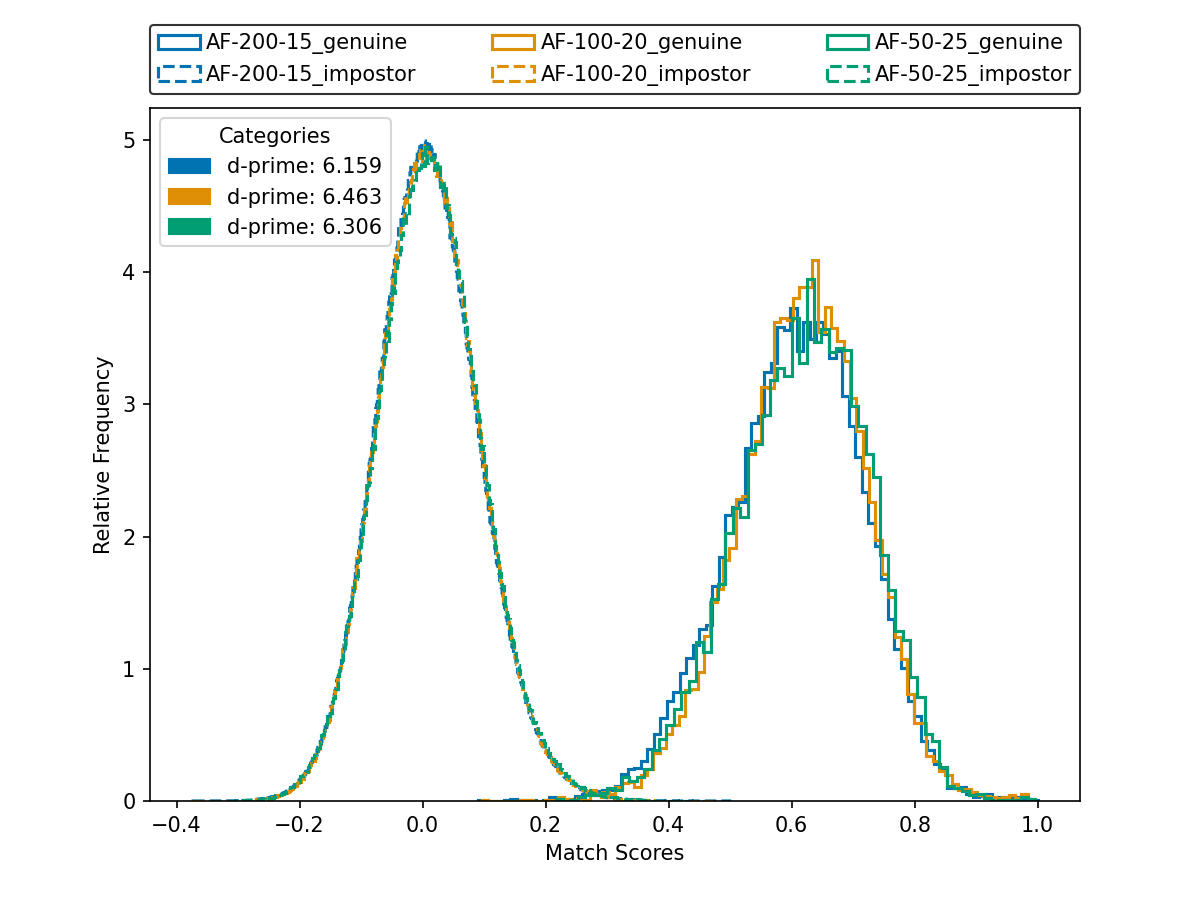}
                
            \end{subfigure}
            \begin{subfigure}[b]{0.24\linewidth}
                \includegraphics[width=\linewidth]{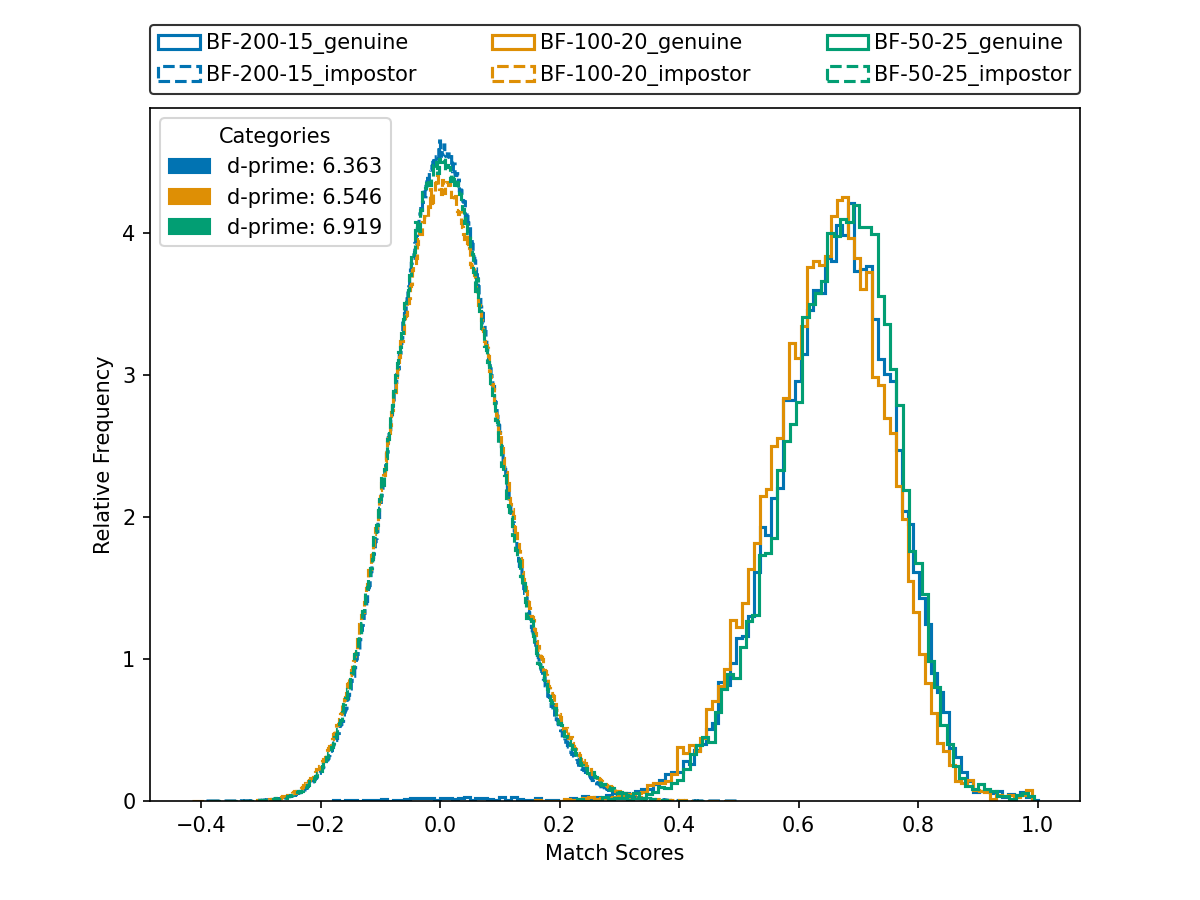}
                
            \end{subfigure}
            \begin{subfigure}[b]{0.24\linewidth}
                \includegraphics[width=\linewidth]{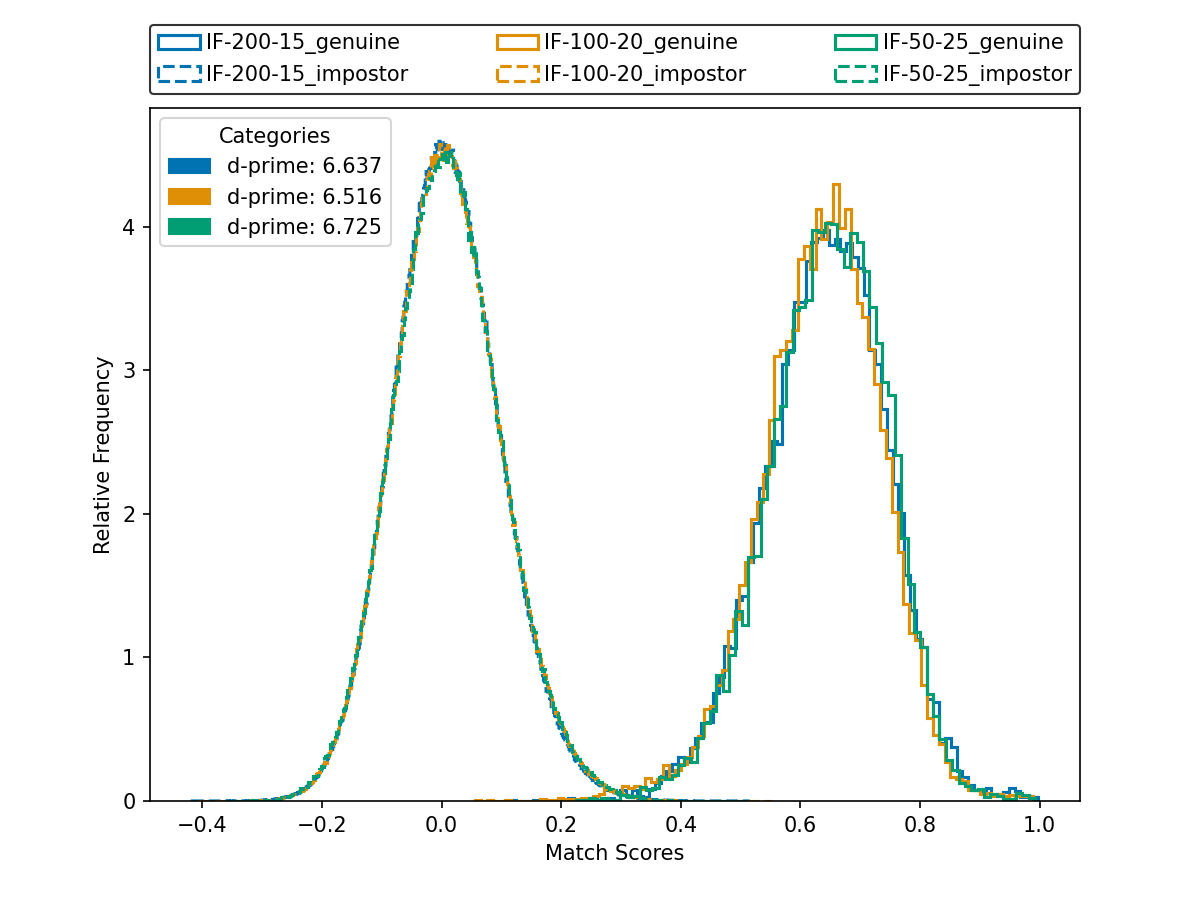}
                
            \end{subfigure}
            \begin{subfigure}[b]{0.24\linewidth}
                \includegraphics[width=\linewidth]{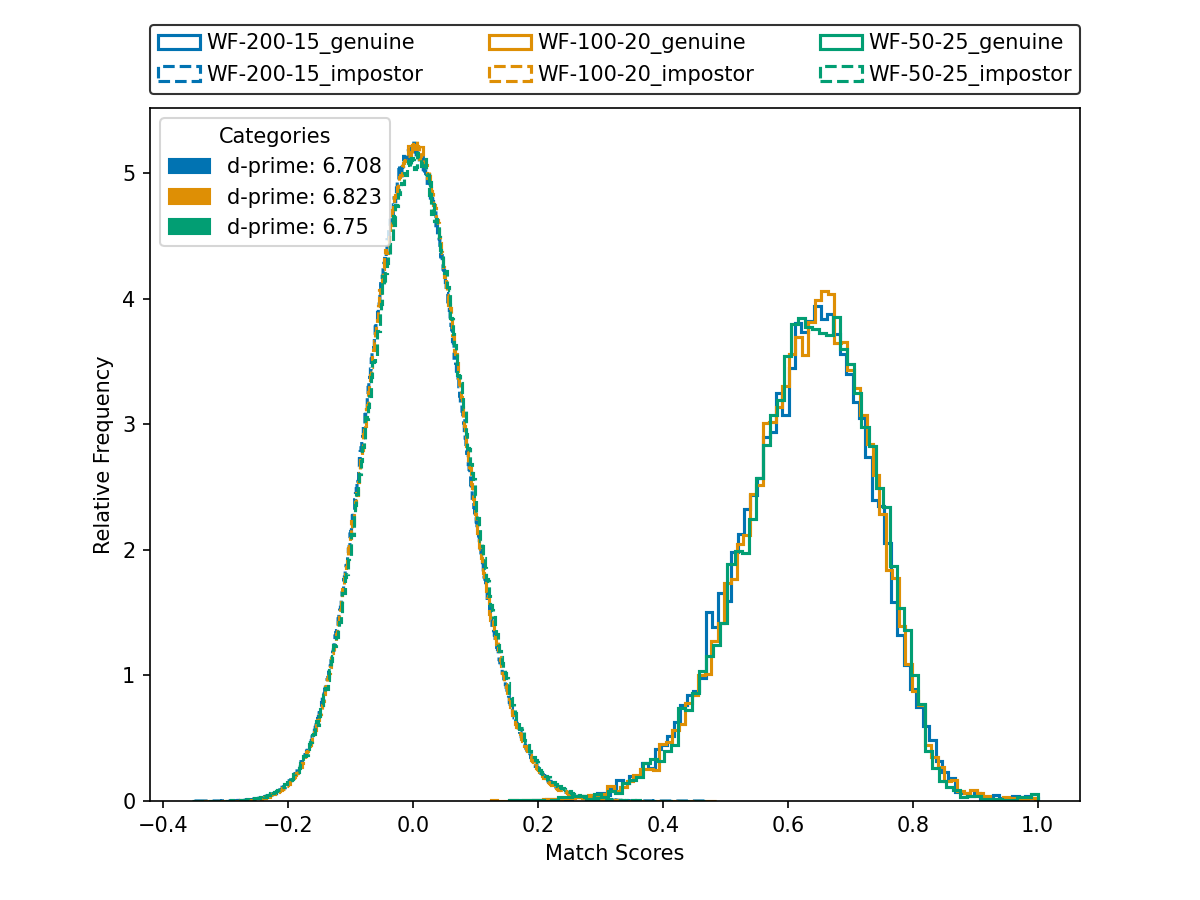}
            \end{subfigure}
        \end{subfigure}
        \begin{subfigure}[b]{1\linewidth}
            \begin{subfigure}[b]{0.24\linewidth}
                \includegraphics[width=\linewidth]{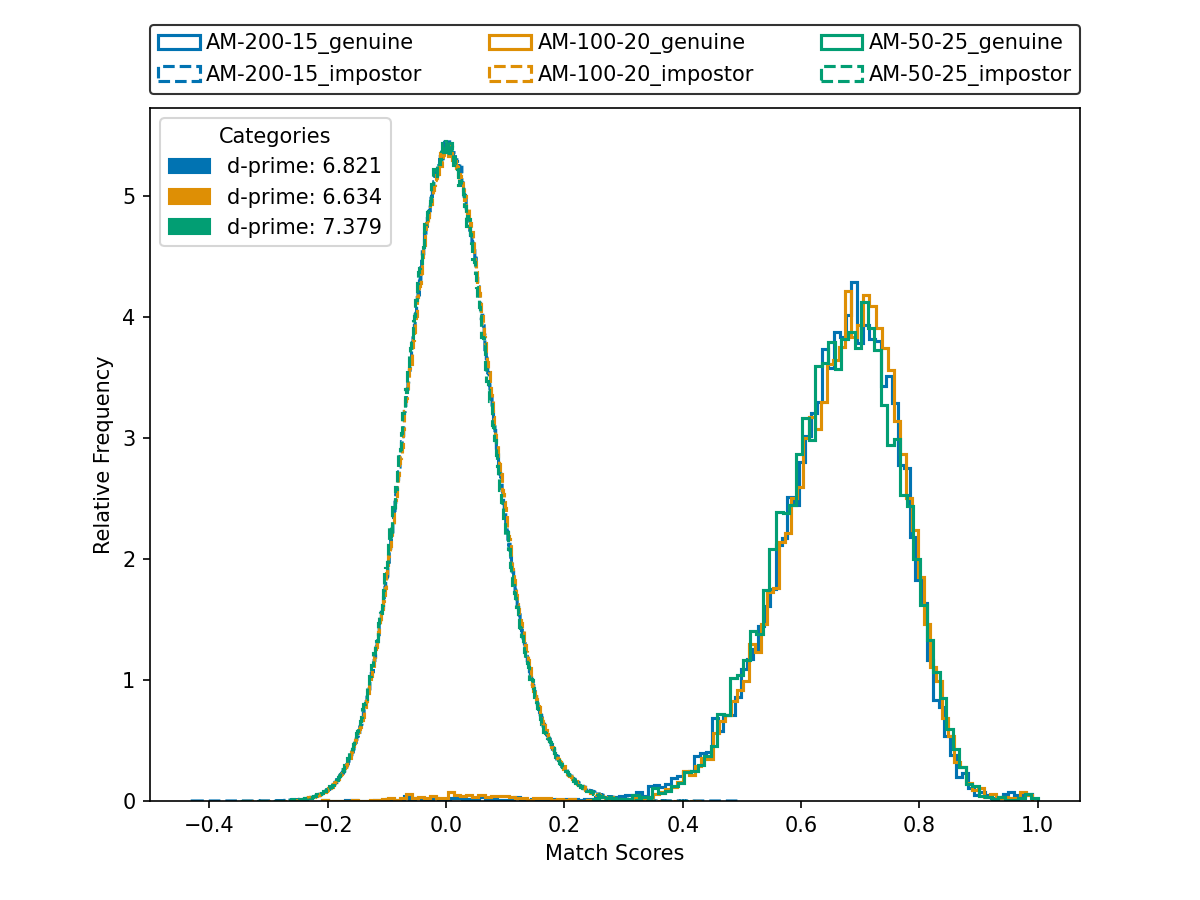}
                
            \end{subfigure}
            \begin{subfigure}[b]{0.24\linewidth}
                \includegraphics[width=\linewidth]{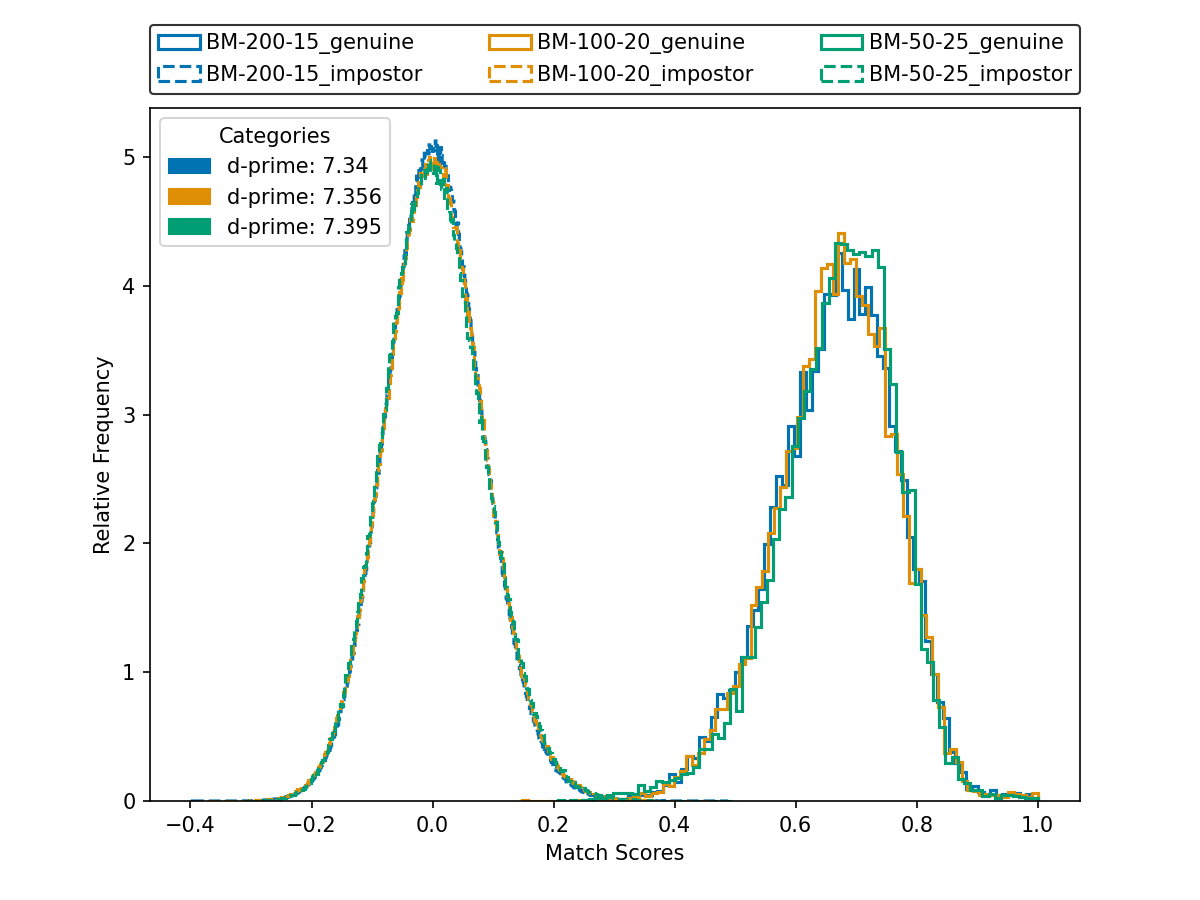}
            \end{subfigure}
            \begin{subfigure}[b]{0.24\linewidth}
                \includegraphics[width=\linewidth]{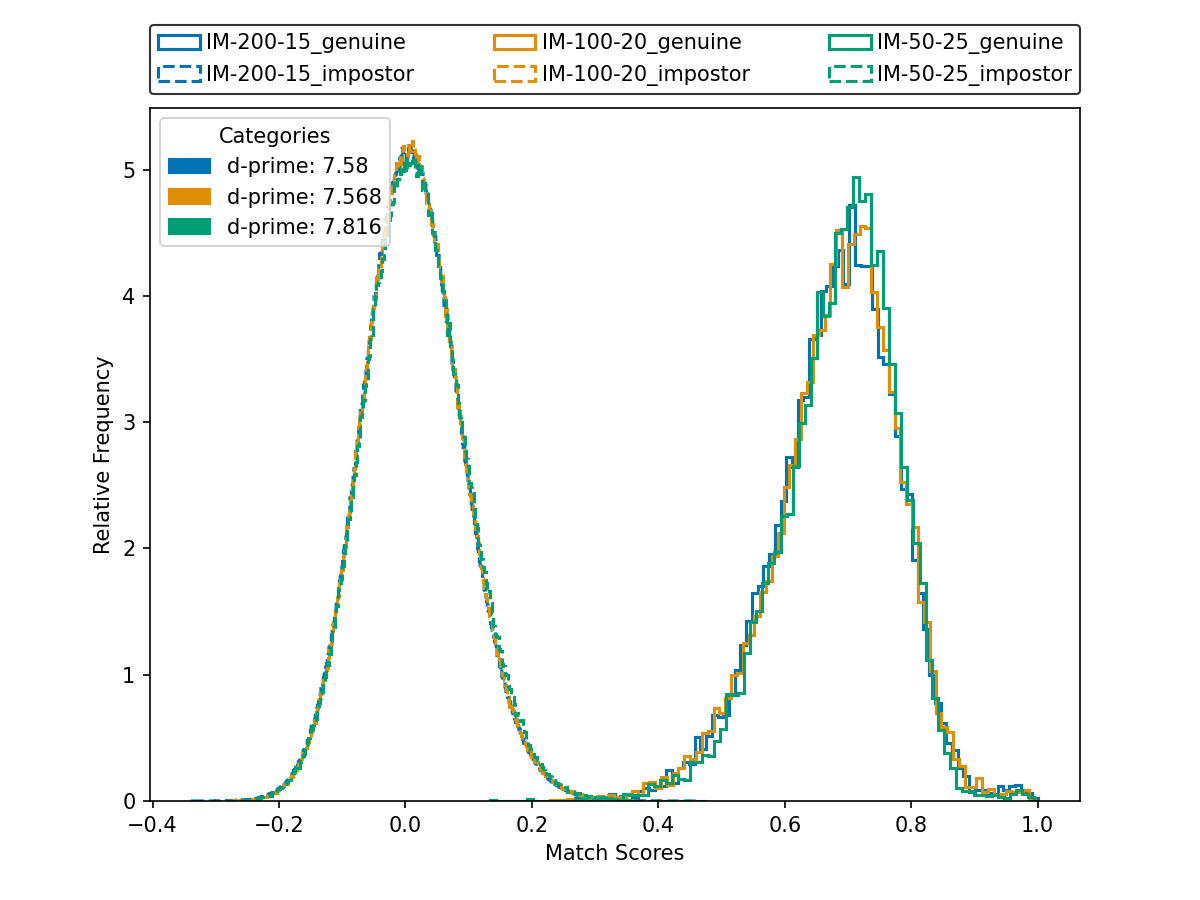}
            \end{subfigure}
            \begin{subfigure}[b]{0.24\linewidth}
                \includegraphics[width=\linewidth]{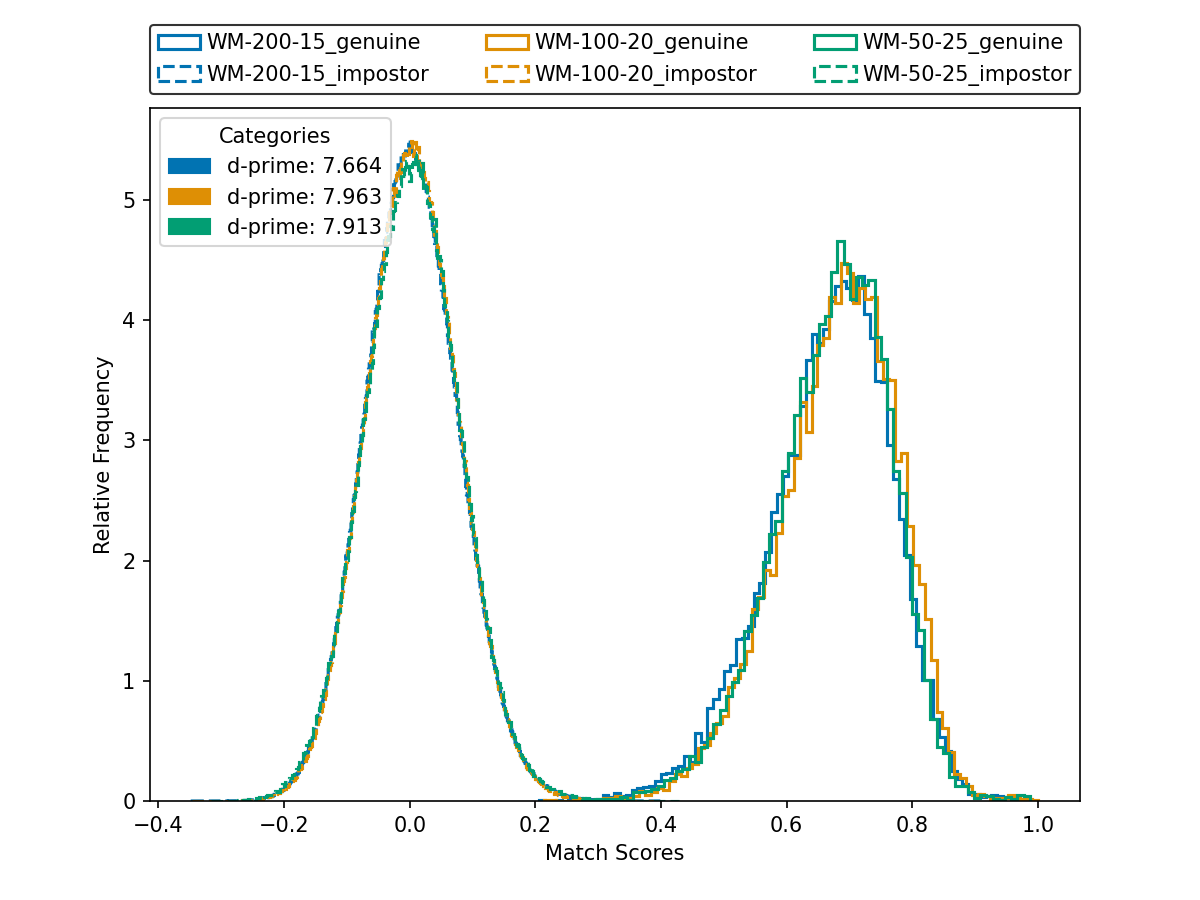}
            \end{subfigure}
        \end{subfigure}
    \end{subfigure}
   \caption{Similarity distributions with the varying number of identities and image per identity for 8 demographics. Top row has distributions of female groups. Bottom row has distribution of male groups. For labels, "AF-200-15" means randomly picking 200 identities with 15 images per identity from Asian Female group.}
\label{fig:id_im_balanced}
\vspace{-2mm}
\end{figure*}

\label{sec:id_im_balanced}
\noindent
\textbf{Is balanced number of IDs and images/ID important for test set?}
Figure~\ref{fig:id_im_balanced} shows the genuine and impostor distributions of eight demographic groups with randomly-selected (200 IDs, 15 images/ID), (100 IDs, 20 images/ID), and (50 IDs, 25 images/ID) from our prepared subset of VGGFace2. Across all groups, there is no significant difference in the impostor or genuine distribution for the different numbers of identities and images. This shows that balancing the number of identities and images across demographic groups in a test set is not relevant to a fair comparison of 1-to-1 matching accuracy. However, the frequency of difficult images (i.e., profile head pose, bad image quality, bad brightness, etc.) and identity noise are major factors impacting the accuracy.

\begin{figure*}[!t]
\centering
    \begin{subfigure}[b]{1\linewidth}
    \centering
        \begin{subfigure}[b]{1\linewidth}
            \begin{subfigure}[b]{0.32\linewidth}
                \includegraphics[width=\linewidth]{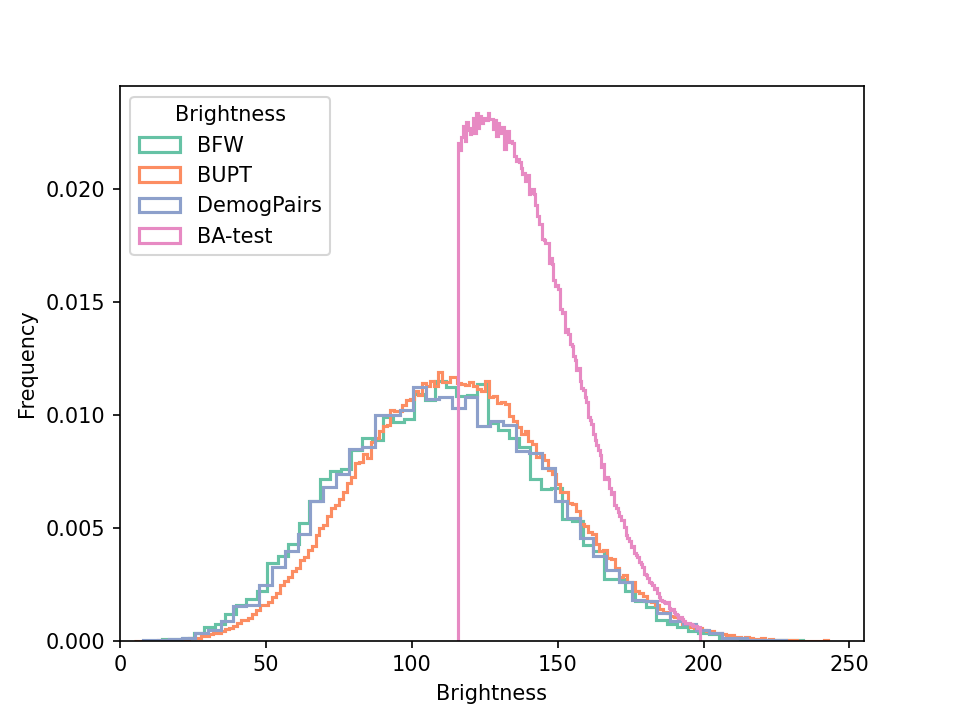}
            \caption{FSB brightness measurement}
            \label{fig:fsb}
            \end{subfigure}
            \begin{subfigure}[b]{0.32\linewidth}
                \includegraphics[width=\linewidth]{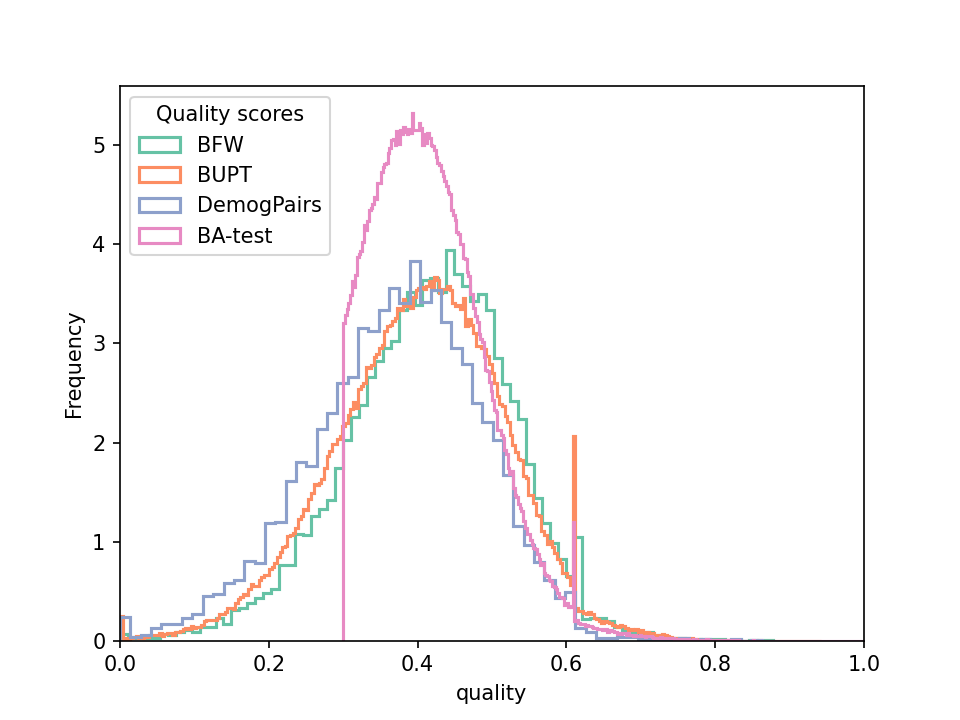}
            \caption{FaceQnet}
            \label{fig:quality_faceqnet}
            \end{subfigure}
            \begin{subfigure}[b]{0.32\linewidth}
                \includegraphics[width=\linewidth]{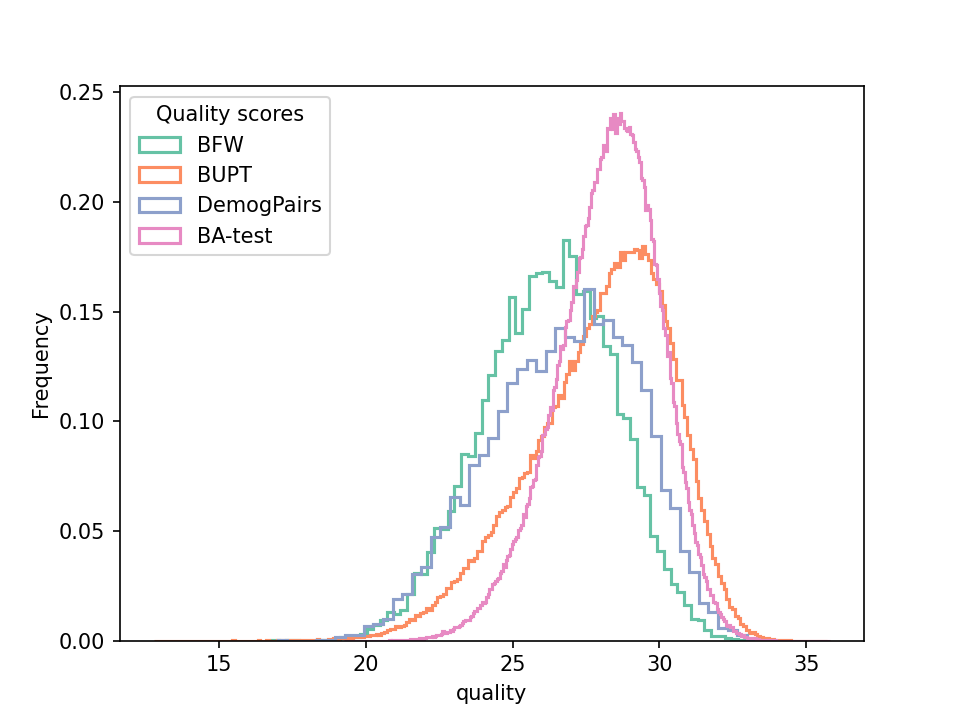}
            \caption{MagFace (quality)}
            \label{fig:magface}
            \end{subfigure}
        \end{subfigure}
    \end{subfigure}
   \caption{Brightness and quality for BFW, DemogPairs, BUPT-BalancedFace, BA-test.}
\label{fig:quality}
\vspace{-3mm}
\end{figure*}
\vspace{-2mm}
\subsection{Analysis of Factors Known to Impact Accuracy}
PIE~\cite{han2013comparative, Sim2003PAMI, gross2010multi} and image quality~\cite{terhorst2023qmagface} are well-known factors that affect the accuracy of facial matching. To reduce their impact, we implement the method in~\cite{albiero2021img2pose} to select the most frontal images. Face Skin Brightness (FSB) metric and the middle-exposed brightness range in~\cite{wu2023face} are used to select images with good brightness. However, unlike controlled acquisition datasets, the face segmentation model does not perform well on in-the-wild images. To ensure accurate FSB measurement, we drop images whose face area predicted by the model is less than 20\% of the image pixels or which have no nose segmentation prediction.
For image quality, FaceQnet \cite{faceqnetv1} and MagFace \cite{magface} are used to select good quality images and cross-check the results. Meanwhile, the distributions of brightness, 3D head pose, and image quality for BFW, BUPT-Balancedface, DemogPairs and the proposed BA-test are shown in Figure~\ref{fig:fsb}, and Figure~\ref{fig:magface} and Figure~\ref{fig:quality_faceqnet}. 

Figure~\ref{fig:fsb} shows that the average brightness of the images in BFW, BUPT-Balancedface, and DemogPairs ranges from less than 10 to over 220. Within this range, both underexposed~\cite{FRVT_2022, wu2023face} and overexposed~\cite{wu2023face} images hurt the similarity of image pairs, reducing reliability of analysis made on these datasets. The head pose in BFW, BUPT-Balancedface, and DemogPairs is not controlled.
As a training set, variation in head pose in BUPT-Balancedface benefits the performance of the face matcher. However, without controlling head pose across demographics in a test set, conclusions about accuracy across demographics may not be reliable.
For image quality, FaceQnet~\cite{faceqnetv1} is trained on the dataset that involves human perception. MagFace~\cite{magface} is trained with a magnitude-based loss function, where the magnitude of the feature represents the quality that the face matcher "thinks" the image is. 
Figure~\ref{fig:quality_faceqnet} and Figure~\ref{fig:magface} show that BA-test has fewer low quality images and less quality variation than the other three datasets, for both quality assessment methods, which should minimize impact of varying image quality on cross-demographic comparisons.

\subsection{Further Analysis On Gender Bias}

\begin{figure*}[t]
\centering
    \begin{subfigure}[b]{1\linewidth}
    \centering

        \begin{subfigure}[b]{1\linewidth}
            \begin{subfigure}[b]{0.24\linewidth}
                \begin{subfigure}[b]{0.48\linewidth}
                \includegraphics[width=\linewidth]{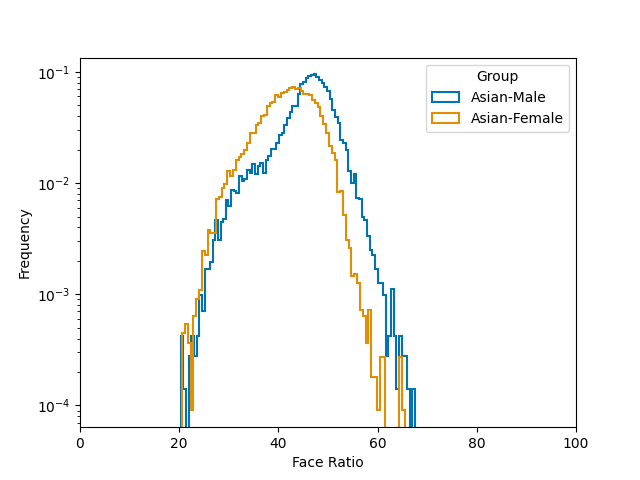}
                \end{subfigure}
                \begin{subfigure}[b]{0.48\linewidth}
                \includegraphics[width=\linewidth]{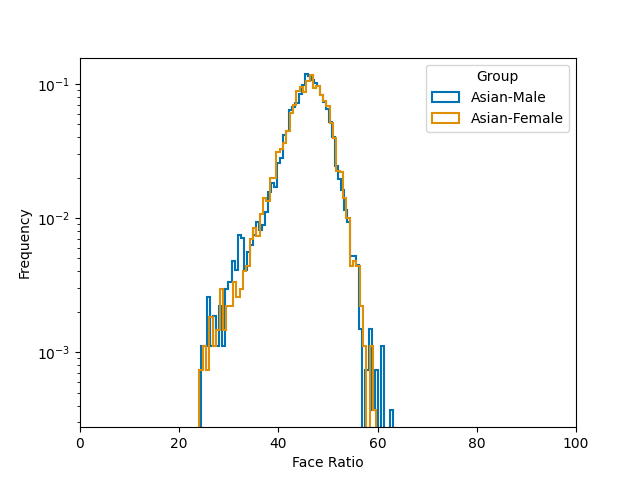}
                \end{subfigure}
            \end{subfigure}
            \begin{subfigure}[b]{0.24\linewidth}
                \begin{subfigure}[b]{0.48\linewidth}
                \includegraphics[width=\linewidth]{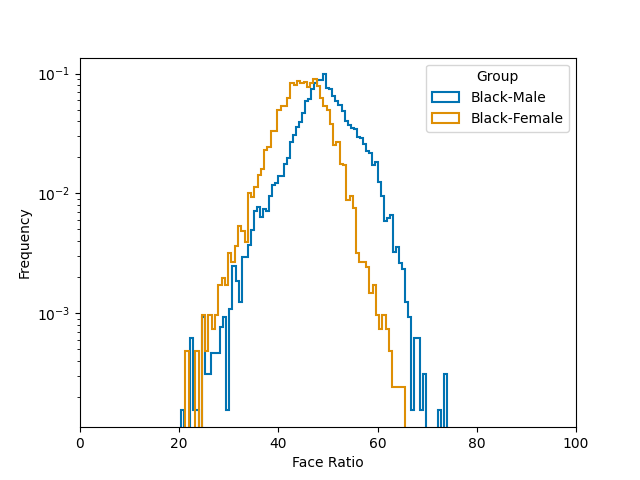}
                \end{subfigure}
                \begin{subfigure}[b]{0.48\linewidth}
                \includegraphics[width=\linewidth]{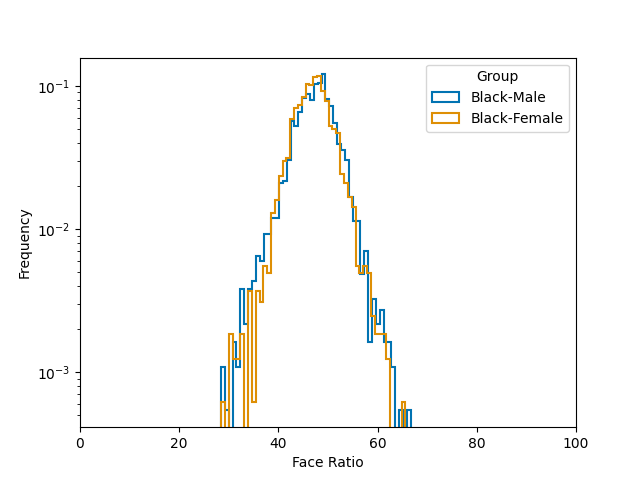}
                \end{subfigure}
            \end{subfigure}
            \begin{subfigure}[b]{0.24\linewidth}
                \begin{subfigure}[b]{0.48\linewidth}
                \includegraphics[width=\linewidth]{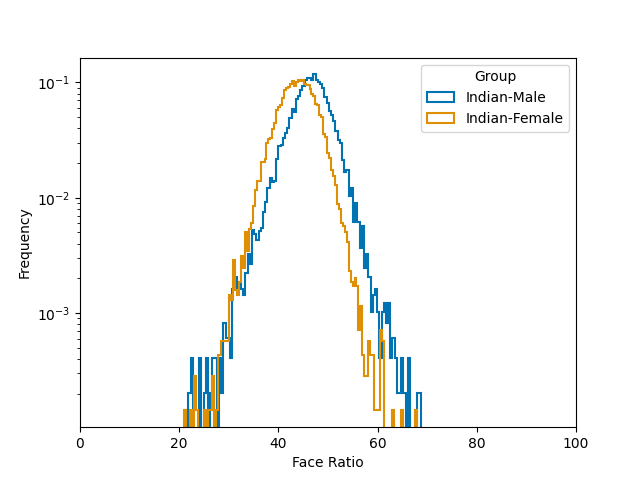}
                \end{subfigure}
                \begin{subfigure}[b]{0.48\linewidth}
                \includegraphics[width=\linewidth]{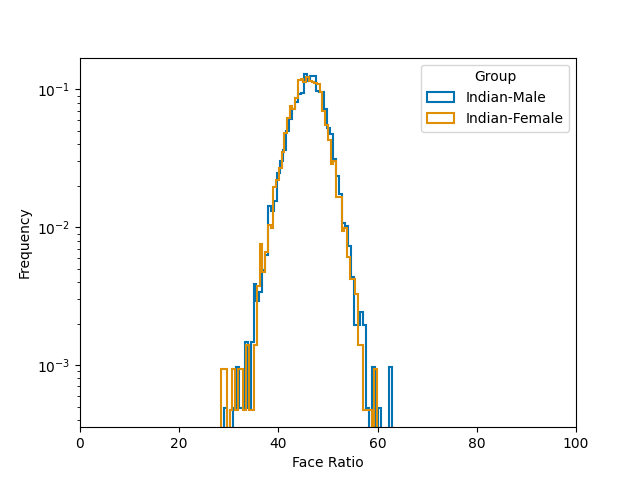}
                \end{subfigure}
            \end{subfigure}
            \begin{subfigure}[b]{0.24\linewidth}
                \begin{subfigure}[b]{0.48\linewidth}
                \includegraphics[width=\linewidth]{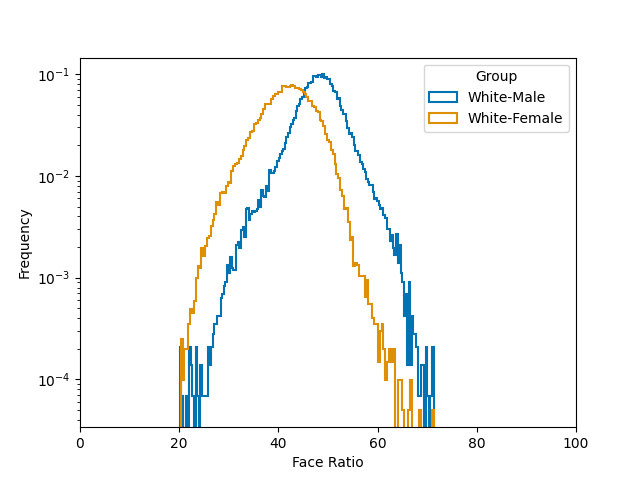}
                \end{subfigure}
                \begin{subfigure}[b]{0.48\linewidth}
                \includegraphics[width=\linewidth]{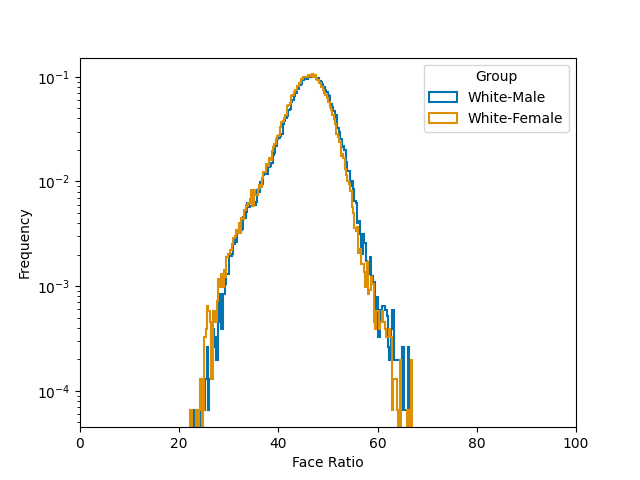}
                \end{subfigure}
            \end{subfigure}
        \end{subfigure}
        
        \begin{subfigure}[b]{1\linewidth}
            \begin{subfigure}[b]{0.24\linewidth}
                \begin{subfigure}[b]{0.48\linewidth}
                \includegraphics[width=\linewidth]{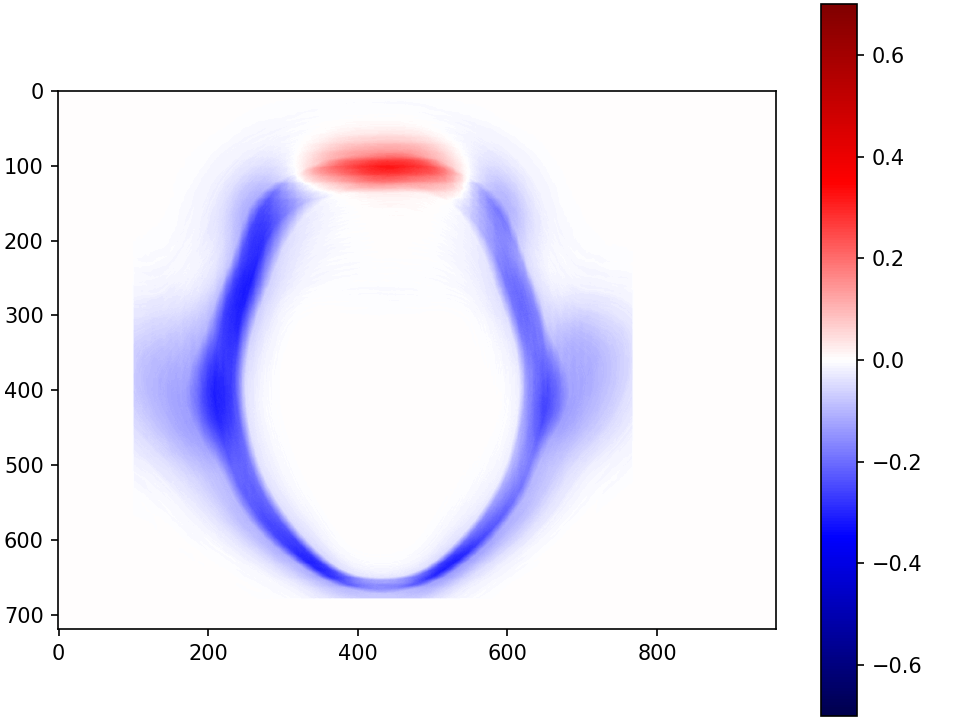}
                \end{subfigure}
                \begin{subfigure}[b]{0.48\linewidth}
                \includegraphics[width=\linewidth]{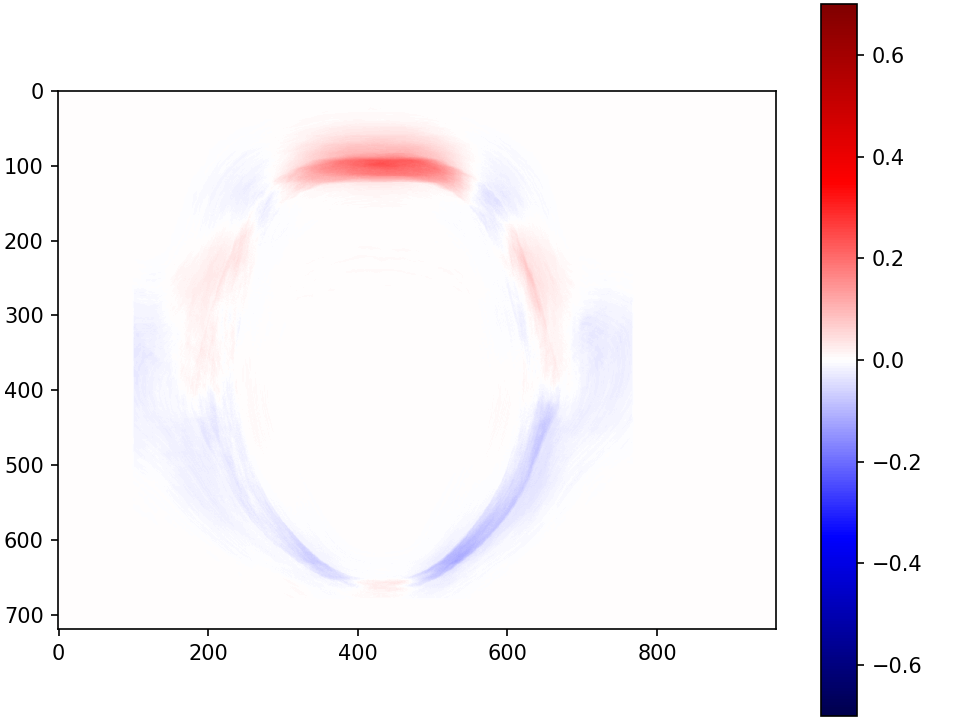}
                \end{subfigure}
            \caption{Asian}
            \end{subfigure}
            \begin{subfigure}[b]{0.24\linewidth}
                \begin{subfigure}[b]{0.48\linewidth}
                \includegraphics[width=\linewidth]{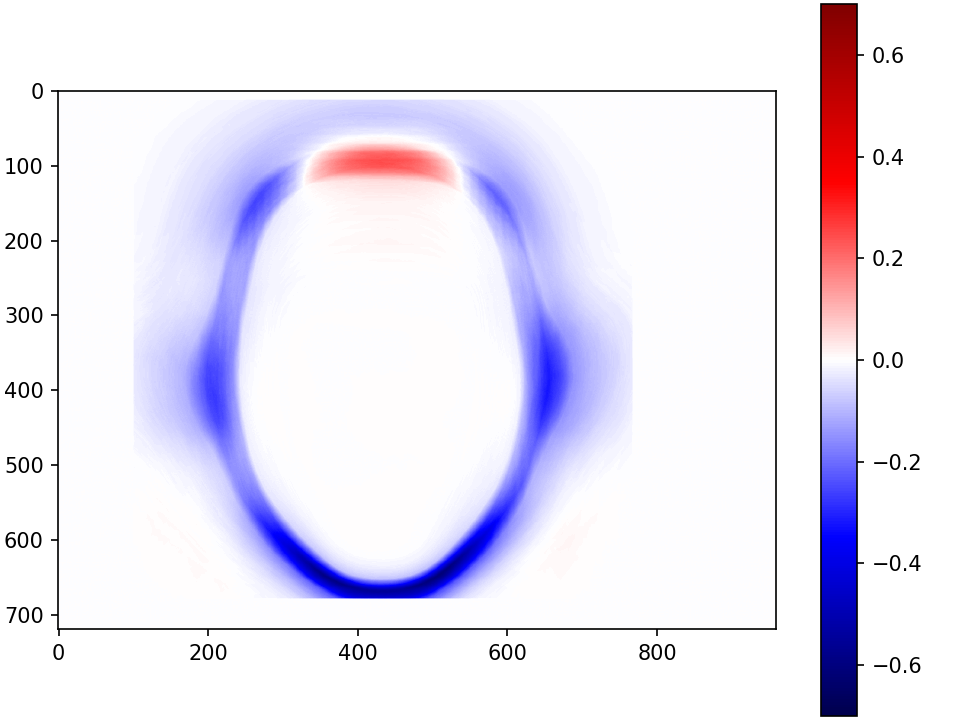}
                \end{subfigure}
                \begin{subfigure}[b]{0.48\linewidth}
                \includegraphics[width=\linewidth]{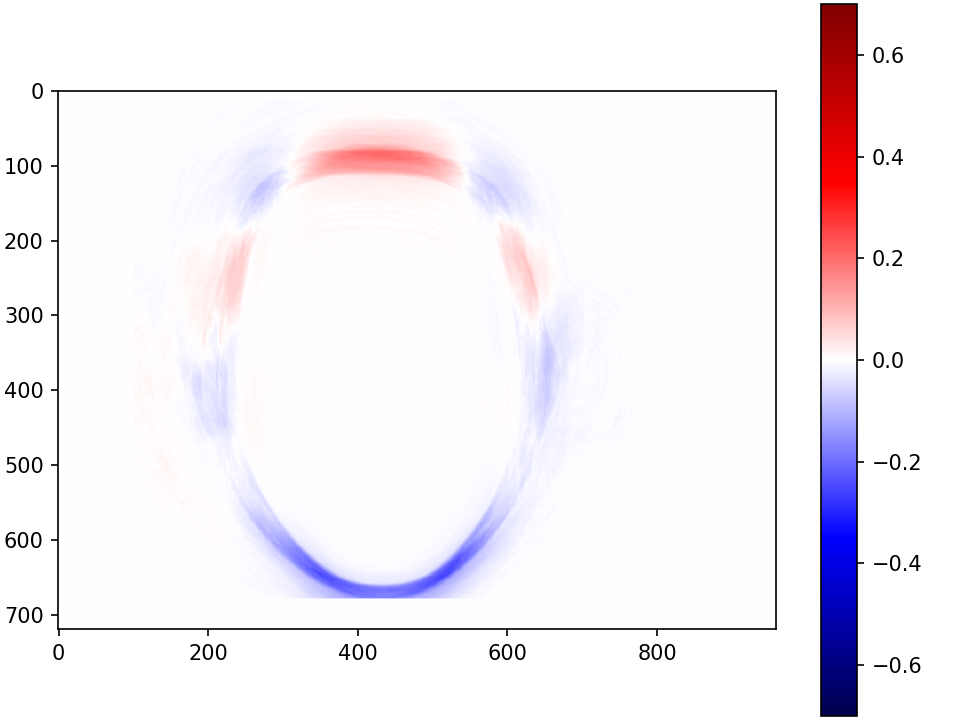}
                \end{subfigure}
            \caption{Black}
            \end{subfigure}
            \begin{subfigure}[b]{0.24\linewidth}
                \begin{subfigure}[b]{0.48\linewidth}
                \includegraphics[width=\linewidth]{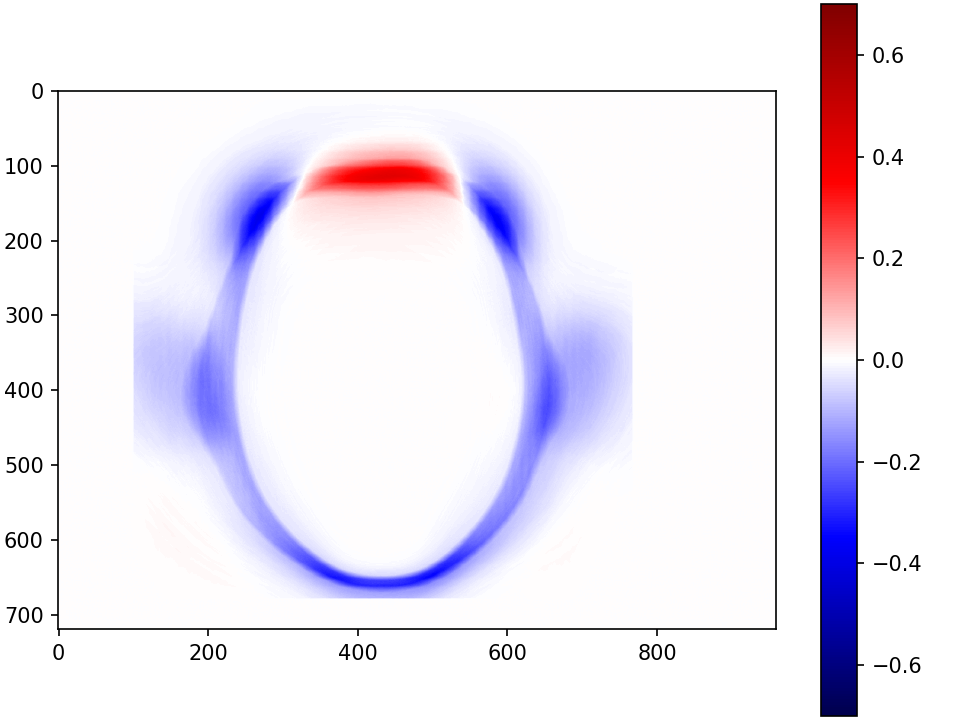}
                \end{subfigure}
                \begin{subfigure}[b]{0.48\linewidth}
                \includegraphics[width=\linewidth]{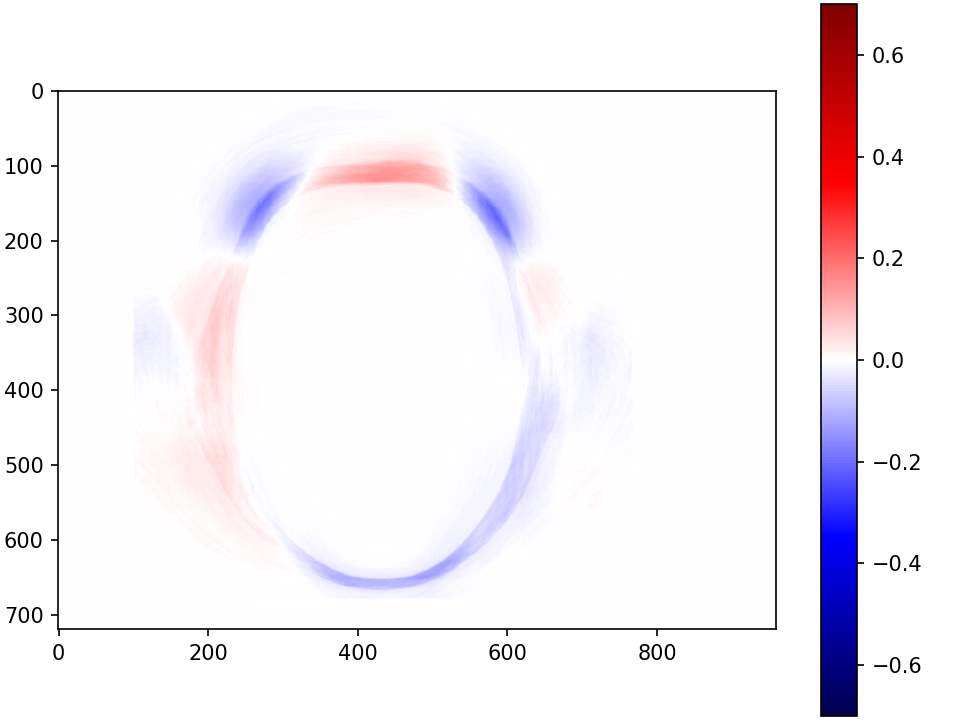}
                \end{subfigure}
            \caption{Indian}
            \end{subfigure}
            \begin{subfigure}[b]{0.24\linewidth}
                \begin{subfigure}[b]{0.48\linewidth}
                \includegraphics[width=\linewidth]{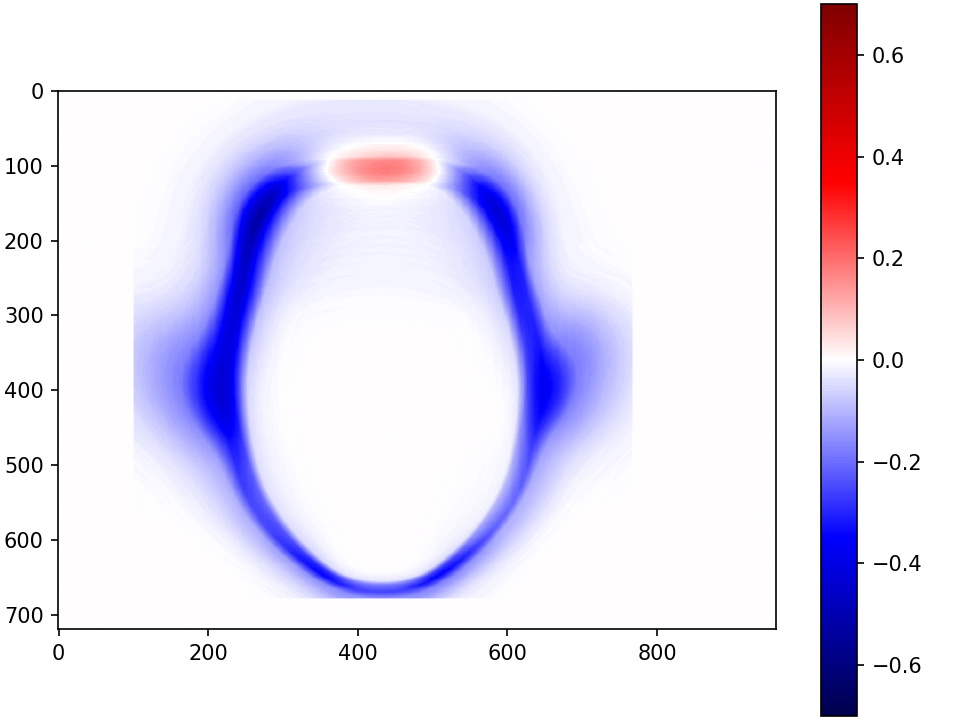}
                \end{subfigure}
                \begin{subfigure}[b]{0.48\linewidth}
                \includegraphics[width=\linewidth]{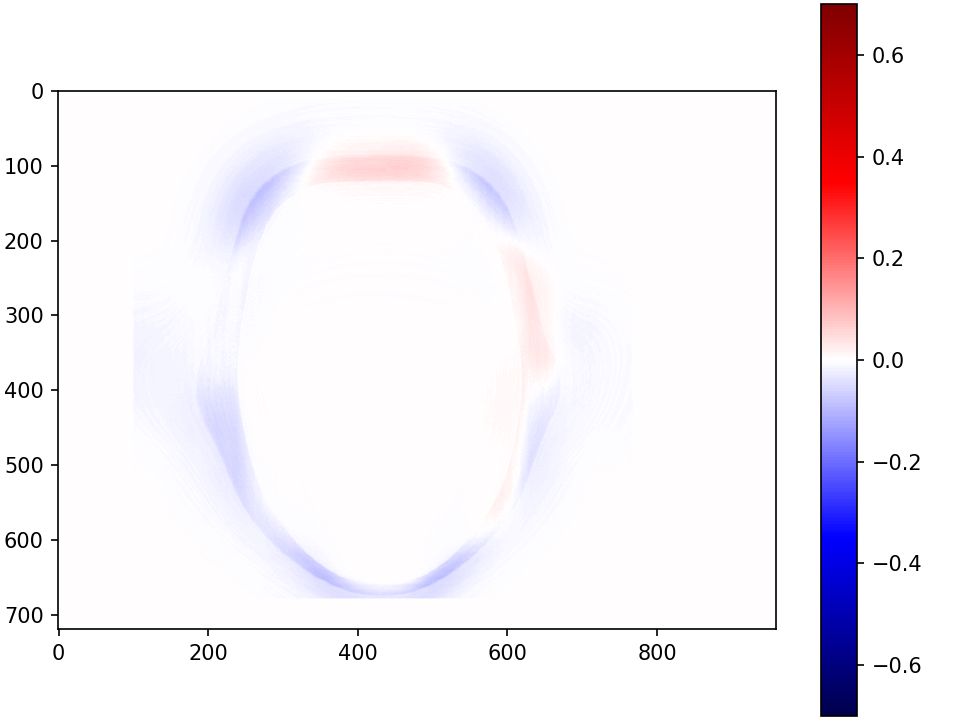}
                \end{subfigure}
            \caption{White}
            \end{subfigure}
        \end{subfigure}
    \end{subfigure}
   \caption{Top: distributions of the visible face area ratio between genders for each race before (left) and after (right) face area balancing. Bottom: heatmaps for difference of average face area between genders before (left) and after (right) balancing  visible face area.}
\label{fig:face_area_balanced}
\vspace{-3mm}
\end{figure*}
Accuracy differences have been reported across gender, age, and race. The first row in Figure~\ref{fig:face_area_balanced} shows that the gender difference, where males have larger visible face area than females, also exists in the proposed dataset. Some researches~\cite{albiero2020face,9650887} report that balancing facial morphology between sexes can reduce the gender gap. Their conclusion is based on a controlled-acquisition dataset, MORPH~\cite{ricanek2006morph}. \textit{Can the observations/conclusions be transferred to in-the-wild data?} is what we investigate in this subsection. Unlike MORPH, images in BA-test are in-the-wild.
We have added balance in brightness, image quality, and head pose. Therefore, our accuracy-factor-balanced version of the in-the-wild dataset may represent images captured under the real-world cases better than MORPH.

\begin{table}[t]
\centering
\small
\begin{tabular}{|c|c|c|c|c|}
\hline
 $\Delta d'$ &
  Asian &
  Black &
  Indian &
  White \\ \hline
 &
  \cellcolor[HTML]{EC9494}0.5336 &
  0.3146 &
  0.5094 &
  \cellcolor[HTML]{EC9494}0.4115 \\ \cline{2-5} 
\multirow{-2}{*}{Original Gen} &
  \cellcolor[HTML]{EC9494}0.5008 &
  0.3393 &
  0.5030 &
  \cellcolor[HTML]{EC9494}0.4164 \\ \hline
 &
  \cellcolor[HTML]{A9D08E}0.2678 &
  \cellcolor[HTML]{A9D08E}0.2447 &
  0.5028 &
  \cellcolor[HTML]{A9D08E}0.284 \\ \cline{2-5} 
\multirow{-2}{*}{Balanced Gen} &
  \cellcolor[HTML]{A9D08E}0.298 &
  \cellcolor[HTML]{A9D08E}0.2824 &
  0.5127 &
  \cellcolor[HTML]{A9D08E}0.2856 \\ \hline
 &
  0.3901 &
  \cellcolor[HTML]{EC9494}0.6813 &
  \cellcolor[HTML]{EC9494}0.8819 &
  0.3719 \\ \cline{2-5} 
\multirow{-2}{*}{Balanced Gen NFH} &
  0.4028 &
  \cellcolor[HTML]{EC9494}0.7288 &
  \cellcolor[HTML]{EC9494}0.8901 &
  0.4095 \\ \hline \hline
 &
  \cellcolor[HTML]{EC9494}0.0211 &
  \cellcolor[HTML]{EC9494}0.1511 &
  \cellcolor[HTML]{A9D08E}0.0745 &
  \cellcolor[HTML]{A9D08E}0.0241 \\ \cline{2-5} 
\multirow{-2}{*}{Original Imp} & \cellcolor[HTML]{A9D08E}0.0125 & \cellcolor[HTML]{EC9494}0.0917 & \cellcolor[HTML]{A9D08E}0.0981 & \cellcolor[HTML]{A9D08E}0.042 \\ \hline
 &
  0.015 &
  0.1262 &
  0.179 &
  0.0398 \\ \cline{2-5} 
\multirow{-2}{*}{Balanced Imp} &
  \cellcolor[HTML]{EC9494}0.0528 &
  \cellcolor[HTML]{A9D08E}0.0505 &
  0.2455 &
  0.0847 \\ \hline
 &
  \cellcolor[HTML]{A9D08E}0.004 &
  \cellcolor[HTML]{A9D08E}0.0377 &
  \cellcolor[HTML]{EC9494}0.2491 &
  \cellcolor[HTML]{EC9494}0.0732 \\ \cline{2-5} 
\multirow{-2}{*}{Balanced Imp NFH} &
  0.0448 &
  0.0733 &
  \cellcolor[HTML]{EC9494}0.304 &
  \cellcolor[HTML]{EC9494}0.1156 \\ \hline
\end{tabular}
\vspace{2mm}
\caption{The gender gap of genuine and impostor distributions for each race before and after balancing the face morphology measured by $\Delta d'$. Gen and Imp are genuine and impostor. NFH means no facial hair. For the $\Delta d'$ of each category, top number is ArcFace model, bottom is MagFace. \textcolor[HTML]{EC9494}{\textbf{Red}} and \textcolor[HTML]{A9D08E}{\textbf{green}} represents the largest and smallest gender gap.}
\label{table:face_area_balanced}
\vspace{-5mm}
\end{table}

We apply the same approach for face area extraction as described in \cite{albiero2020face}. (Note that images with a predicted face area of less than 20\% were discarded earlier.) As depicted in Figure~\ref{fig:face_area_balanced}, males generally have a larger visible face area than females. The heatmaps were generated by calculating the difference between the average face areas of females and males, with blue representing pixels more frequently labeled as face for males than females, and red indicating the opposite.

Since  images are aligned based on eye position, our findings suggest females have greater distance between the top of their face and their eyes, while males have a longer distance between the face center and other regions. Previous research attributes this  to gendered hairstyles, but this does not explain the more pronounced distance from face center to the jawline in males. We speculate this could be due to facial hair, as the face segmentation model classifies facial hair as ``skin'', causing a bearded chin to appear longer.

To balance face areas, we selected image pairs with a face area intersection-over-union (IoU) greater than 0.9. The face ratio distributions and heatmaps reveal a significant reduction in face area differences. Table~\ref{table:face_area_balanced} presents the gender gap, measured by $\Delta d'$, before and after balancing face areas. For genuine pairs, the gender gap decreases by an average of 24.11\% after balancing face areas for both face matchers, with the largest decrease of 45.3\% observed in Asians, and a slight increase for Indians. For impostor pairs, the gender gap decreases by 27.22\% for Blacks, but increases by 151.5\% for Indians and 88.35\% for Whites on average. The two face matchers exhibit different trends for Asians.

Studies by \cite{wu2023logical} and \cite{bhatta2023gender} report that facial hair significantly impacts face recognition accuracy. By removing all male samples with facial hair and balancing face areas (shown in the third and sixth rows of Table~\ref{table:face_area_balanced}), we observe that gender gaps for genuine pairs increase compared to the balanced version without considering facial hair. For Blacks and Indians, the gender gap averages a 116\% and 75\% increase, respectively, compared to the original gender gap. For impostor pairs, the gender gap for Asians and Blacks is similar to or smaller than the other two groups. However, for Indians and Whites, the gender gap averages a 222\% and 189\% increase, respectively, compared to the original gender gap.

In conclusion, balancing facial morphology decreases the gender gap for genuine pairs across all four races. For impostor pairs, this approach reduces the gender gap for Blacks but increases the gap for Indians and Whites. It is important to note that the two face matchers perform differently for Asian impostor pairs, suggesting that different matchers may be sensitive to varying factors. Moreover, facial hair have a significant impact on gender accuracy disparity and more fine-grain analysis on facial hair is our future work.

\section{Bias-Aware Test Dataset}
\begin{table}[]
\centering
\small
\begin{tabular}{|ccccccc|}
\hline
\multicolumn{7}{|c|}{Statistic information}                                                                                                                 \\ \hline
\multicolumn{1}{|c|}{Datasets} &
  \multicolumn{1}{c|}{Data sources} &
  \multicolumn{1}{c|}{IDs} &
  \multicolumn{1}{c|}{Images} &
  \multicolumn{1}{c|}{Subgroups} &
  \multicolumn{1}{c|}{Age} &
  ID denoise \\ \hline
\multicolumn{1}{|c|}{DemogPairs} &
  \multicolumn{1}{c|}{\cite{yi2014learning, parkhi2015deep, vggface2}} &
  \multicolumn{1}{c|}{600} &
  \multicolumn{1}{c|}{10,800} &
  \multicolumn{1}{c|}{6} &
  \multicolumn{1}{c|}{\xmark} &
  \multicolumn{1}{c|}{\xmark} \\ \hline
\multicolumn{1}{|c|}{RFW} &
  \multicolumn{1}{c|}{\cite{guo2016ms}} &
  \multicolumn{1}{c|}{12,000} &
  \multicolumn{1}{c|}{80,000} &
  \multicolumn{1}{c|}{4} &
  \multicolumn{1}{c|}{\xmark} &
  \multicolumn{1}{c|}{\xmark} \\ \hline
\multicolumn{1}{|c|}{BFW} &
  \multicolumn{1}{c|}{\cite{vggface2}} &
  \multicolumn{1}{c|}{800} &
  \multicolumn{1}{c|}{20,000} &
  \multicolumn{1}{c|}{8} &
  \multicolumn{1}{c|}{\xmark} &
  \multicolumn{1}{c|}{\xmark} \\ \hline
\multicolumn{1}{|c|}{BA-test (ours)} &
  \multicolumn{1}{c|}{\cite{vggface2}} &
  \multicolumn{1}{c|}{8,321} &
  \multicolumn{1}{c|}{177,227} &
  \multicolumn{1}{c|}{8} &
  \multicolumn{1}{c|}{2} &
  \multicolumn{1}{c|}{\cmark} \\ \hline \hline
\multicolumn{7}{|c|}{Balanced factors}                                                                                                                   \\ \hline
\multicolumn{1}{|c|}{Datasets} &
  \multicolumn{1}{c|}{Head pose} &
  \multicolumn{1}{c|}{Race} &
  \multicolumn{1}{c|}{Quality} &
  \multicolumn{1}{c|}{Brightness} &
  \multicolumn{1}{c|}{ID} &
  Gender\\ \hline
\multicolumn{1}{|c|}{DemogPairs} & \multicolumn{1}{c|}{\xmark} & \multicolumn{1}{c|}{\cmark} & \multicolumn{1}{c|}{\xmark} & \multicolumn{1}{c|}{\xmark} & \multicolumn{1}{c|}{\cmark} & \multicolumn{1}{c|}{\cmark} \\ \hline
\multicolumn{1}{|c|}{RFW}        & \multicolumn{1}{c|}{\xmark} & \multicolumn{1}{c|}{\cmark} & \multicolumn{1}{c|}{\xmark} & \multicolumn{1}{c|}{\xmark} & \multicolumn{1}{c|}{\xmark} & \multicolumn{1}{c|}{\xmark} \\ \hline
\multicolumn{1}{|c|}{BFW}        & \multicolumn{1}{c|}{\xmark} & \multicolumn{1}{c|}{\cmark} & \multicolumn{1}{c|}{\xmark} & \multicolumn{1}{c|}{\xmark} & \multicolumn{1}{c|}{\cmark} & \multicolumn{1}{c|}{\cmark} \\ \hline
\multicolumn{1}{|c|}{BA-test (ours)}    & \multicolumn{1}{c|}{\cmark} & \multicolumn{1}{c|}{\xmark} & \multicolumn{1}{c|}{\cmark} & \multicolumn{1}{c|}{\cmark} & \multicolumn{1}{c|}{\xmark} & \multicolumn{1}{c|}{\xmark} \\ \hline
\end{tabular}
\vspace{2mm}
\caption{Existing demographically-balanced test datasets. Upper table gives source of data, number of identities, images, demographic groups, ages, and whether identity labels have been denoised.  Bottom table shows factors balanced in each dataset.}
\end{table}
\begin{table}[t]
    \centering
    \begin{tabular}{|c|c|c|c|c|c|c|c|c|}
    \hline
    FMR     & BM    & BF     & IM     & IF     & WM    & WF     & AM    & AF    \\ \hline
    BFW     & 0.0161& 0.0357 & 0.0323 & 0.0334 & 0.001& 0.0080 & 0.0225& 0.0533 \\ \hline
    VGGFace2    & 0.0059  & 0.0314   & 0.0174   & 0.0273   & 0.001  & 0.0026  & 0.0143  & 0.0267      \\ \hline
    BA-test & \textbf{0.005} & \textbf{0.0302} & \textbf{0.0141} & \textbf{0.0251} & 0.001 & \textbf{0.0024} & \textbf{0.0100} & \textbf{0.0160} \\ \hline
    \end{tabular}
    \vspace{1mm}
    \captionof{table}{False match rate comparisons with 1-in-10,000 threshold value of White Males in each dataset. Note that both BFW and BA-test are purely assembled on VGGFace2.}
    \vspace{-5mm}
    \label{table:fmr}
\end{table}
The proposed dataset, BA-test, contains 177K images from 8K identities, which is larger than the existing related datasets. Since it is assembled from a single data source, the reliability of identity classification should be higher than that of multiple sources. It has 8 demographic groups with sufficient images per group: 
45,642 images of 3,631 White males, 
53,245 images of 2,865 White females, 
13,311 images of 288 Asian males,
19,454 images of 277 Asian females, 
10,610 images of 577 Black males,
6,190 images of 188 Black females,
11,091 images of 244 Indian males,
17,684 images of 251 Indian females.
In addition, the images are classified in three age groups - Young, Middle\_Aged, Senior. However, due to the small number of seniors in the original dataset (e.g., 77 senior black women, 56 senior Asian females), only Young and Middle\_Aged images are selected. 

As discussion in Section~\ref{sec:id_im_balanced}, the number of identities and images across demographic groups is not necessary to be balanced for face verification, so this imbalanced version is proposed. However, for face identification (1-to-many matching), the number of images and identities do affect accuracy~\cite{drozdowski2021watchlist, grother2004models}. In the BA-test, there are 7,896 identities that have more than 2 images, 5,335 identities that have more than 10 images, and 2,969 identities that have more than 20 images. Hence, BA-test can potentially be used for face identification analysis. Moreover, the identity noise, samples in Figure~\ref{fig:noise_samples}, has been reduced. Head pose, brightness, and image quality are balanced in our dataset, but not in the others. Therefore, conclusions about cross-demographic accuracy difference based on our dataset should be more reliable.

\subsection{Racial Accuracy Disparity}
To illustrate the advantage of this work, we measures the false match rate of each demographic groups, with the 1-in-10,000 threshold value of White Males in the BFW, VGGFace2, and BA-test datasets. Since BFW and BA-test are assembled based on VGGFace2, it is fair to compare the results. However, due to the large amount of images VGGFace2, it is not worth doing to measure the similarity of image pairs across the whole dataset. We first run FairFace to get the demographic labels of VGGFace2, then randomly picking the same amount of identities and images as that of the BA-test dataset. Table~\ref{table:fmr} shows that BA-test has the smallest FMRs for all demographic groups and the smallest gender gap across races.

\subsection{Benchmarks on Bias}

\label{sec:Benchmarks}
\begin{table*}[t]
\centering
\small
\begin{tabular}{ccc||ccc||ccc}
Loss &
  Model &
  Train &
  AF &
  AM &
  diff. &
  WF &
  WM &
  diff. \\ \hline
MagFace &
  r50 &
  Mv2 &
  \cellcolor[HTML]{EC9494}65.00 &
  \cellcolor[HTML]{EC9494}79.78 &
  \cellcolor[HTML]{EC9494}14.78 &
  76.67 &
  86.22 &
  9.56 \\
MagFace &
  r100 &
  Mv2 &
  \cellcolor[HTML]{EC9494}81.56 &
  \cellcolor[HTML]{EC9494}94.44 &
  \cellcolor[HTML]{EC9494}12.89 &
  89.44 &
  \cellcolor[HTML]{A9D08E}96.56 &
  7.11 \\
ArcFace &
  r100 &
  Mv2 &
  \cellcolor[HTML]{EC9494}81.56 &
  \cellcolor[HTML]{EC9494}93.11 &
  \cellcolor[HTML]{EC9494}11.56 &
  90.11 &
  \cellcolor[HTML]{A9D08E}97.11 &
  7.00 \\
ArcFace &
  r50 &
  Glint &
  \cellcolor[HTML]{EC9494}81.22 &
  \cellcolor[HTML]{EC9494}93.00 &
  \cellcolor[HTML]{EC9494}11.78 &
  92.67 &
  \cellcolor[HTML]{A9D08E}95.78 &
  3.11 \\
ArcFace &
  r100 &
  Glint &
  \cellcolor[HTML]{EC9494}90.00 &
  \cellcolor[HTML]{EC9494}96.78 &
  \cellcolor[HTML]{EC9494}6.78 &
  95.78 &
  \cellcolor[HTML]{A9D08E}98.78 &
  3.00 \\ \hline
Loss &
   &
   &
  BF &
  BM &
  diff. &
  IF &
  IM &
  diff. \\ \hline
MagFace &
  r50 &
  Mv2 &
  85.56 &
  86.78 &
  1.22 &
  \cellcolor[HTML]{A9D08E}86.78 &
  \cellcolor[HTML]{A9D08E}90.78 &
  4.00 \\
MagFace &
  r100 &
  Mv2 &
  91.00 &
  94.22 &
  3.22 &
  \cellcolor[HTML]{A9D08E}96.00 &
  96.11 &
  0.11 \\
ArcFace &
  r100 &
  Mv2 &
  91.56 &
  94.11 &
  2.56 &
  \cellcolor[HTML]{A9D08E}94.56 &
  95.56 &
  1.00 \\
ArcFace &
  r50 &
  Glint &
  93.44 &
  93.78 &
  0.33 &
  \cellcolor[HTML]{A9D08E}94.89 &
  93.67 &
  -1.22 \\
ArcFace &
  r100 &
  Glint &
  98.00 &
  97.67 &
  -0.33 &
  \cellcolor[HTML]{A9D08E}98.56 &
  97.22 &
  -1.33
\end{tabular}
\vspace{2mm}
\caption{True positive rates (\%) with a false match rate of $10^{-5}$ and the best (\textbf{\textcolor[HTML]{A9D08E}{green}}) and worst (\textbf{\textcolor[HTML]{EC9494}{red}}) accuracy for each face matcher across eight demographic groups. diff. is the highest TPR - the lowest TPR in each block. Mv2 and Glint are MS1MV2 and Glint360K.}
\vspace{-5mm}
\label{table:demographic}
\end{table*}

BA-test is over-large as a testing benchmars. Since image quality is balanced in a good range, it is not a challenge for state-of-the-art face matchers. However, this dataset is a good indicator for measuring how biased models may be on gender, age, and race. To pick challenging samples, we first normalized the two quality measurements into $[0,1]$ by Min-Max normalization where $Q$ is a image quality vector of BA-test. After aggregating the quality values, for each demographic group, we used the image quality value at 50\% percentile $Q_{50th}$ to randomly pick 90 subjects with 5 images of relatively poor quality ($<Q_{50th}$) images per subject. Consequently, there are 3,600 images from 720 identities, where there are 2,465 images in Young, 1,135 images in Middle-Aged, 1,800 females, and 1,800 males. We evaluated pre-trained face matchers from~\cite{arcface,magface,an2021partial} with  ResNet50 and ResNet100 backbones trained with MS1MV2 and Glint360K.

To better compare  accuracy in age, gender and race, we use the 1-in-100K false match rate (FMR) of all the impostor similarities as a decision threshold to calculate the true positive rate (TPR). 
A general conclusion for age, gender and race is that accuracy disparity exists in all five matchers even though they are different algorithms and have different accuracy. For age, the TPR of Middle-Aged is 4.98\% higher than Young. Males have 4.86\% higher TPR than females. For race, matchers perform best on Indian and worst on Asian, the difference on average is 8.77\%. White and Black do not have a general conclusion on accuracy across the matchers. Results suggest that the largest difference in cross-demographic accuracy is based on race.

Table~\ref{table:demographic} shows the TPR of each demographic with the same 1-in-100K FMR threshold. For gender bias, matchers are biased Asian $>$ White $>$ Black $>$ Indian, where the average gender gap for Asian is 11.6\%, for White is 6\%, for Black is 1.4\%, and 0.5\% for Indian. Results show that gender gap is smaller for Indian and Black, but larger for Asian and White. Furthermore, the difference between best (White Male) and worst (Asian Female) accuracy is 15\% on average. Again, test samples are balanced in quality; thus, the difference in accuracy reflects the bias of the models caused by gender, race, and age. Therefore, even though current matchers have high accuracy, demographic differences are still an issue.
\vspace{-3mm}

\section{Conclusion}
\label{sec:conclusion}

This work demonstrates that balancing the number of identities and images per identity is insufficient to address bias in 1-to-1 matching. Instead, factors such as head pose, brightness, image quality, and gendered characteristics play critical roles in understanding bias.

We propose a bias-aware toolkit for assembling datasets and creating bias-aware test sets (BA-test). This test set, with more identities and images, enables researchers to draw reliable conclusions about the sources of bias in real-world scenarios.

We introduce a face recognition bias benchmark dataset and evaluate three state-of-the-art models, revealing that age gap, gender gap, and demographic accuracy disparity persist.

Future research includes exploring additional bias-contributing factors, examining their impact in face identification, and developing algorithms to mitigate bias in face recognition.
\vspace{-3mm}

\section{Acknowledgement}

Thanks to the help from Dr. V\'itor Albiero. This research is based upon work supported in part by the Office of the Director of National Intelligence (ODNI), Intelligence Advanced Research Projects Activity (IARPA), via \textit{2022-21102100003}. The views and conclusions contained herein are those of the authors and should not be interpreted as necessarily representing the official policies, either expressed or implied, of ODNI, IARPA, or the U.S. Government. The U.S. Government is authorized to reproduce and distribute reprints for governmental purposes notwithstanding any copyright annotation therein.

\bibliography{egbib}

\begin{thebibliography}{62}
\providecommand{\natexlab}[1]{#1}
\providecommand{\url}[1]{\texttt{#1}}
\expandafter\ifx\csname urlstyle\endcsname\relax
  \providecommand{\doi}[1]{doi: #1}\else
  \providecommand{\doi}{doi: \begingroup \urlstyle{rm}\Url}\fi

\bibitem[FQN()]{FQN_github}
https://github.com/javier-hernandezo/FaceQnet, last accessed on April 2023.

\bibitem[bis()]{bisenet_github}
https://github.com/zllrunning/face-parsing.PyTorch, last accessed on February
  2021.

\bibitem[Albiero and Bowyer(2020)]{albiero2020face}
V{\'\i}tor Albiero and Kevin~W Bowyer.
\newblock Is face recognition sexist? no, gendered hairstyles and biology are.
\newblock In \emph{British Machine Vision Conference (BMVC)}, 2020.

\bibitem[Albiero et~al.(2020{\natexlab{a}})Albiero, Bowyer, Vangara, and
  King]{albiero2020does}
V{\'\i}tor Albiero, Kevin Bowyer, Kushal Vangara, and Michael King.
\newblock Does face recognition accuracy get better with age? deep face
  matchers say no.
\newblock In \emph{Proceedings of the IEEE/CVF Winter Conference on
  Applications of Computer Vision}, pages 261--269, 2020{\natexlab{a}}.

\bibitem[Albiero et~al.(2020{\natexlab{b}})Albiero, KS, Vangara, Zhang, King,
  and Bowyer]{albiero2020analysis}
Vitor Albiero, Krishnapriya KS, Kushal Vangara, Kai Zhang, Michael~C King, and
  Kevin~W Bowyer.
\newblock Analysis of gender inequality in face recognition accuracy.
\newblock In \emph{Proceedings of the IEEE/CVF Winter Conference on
  Applications of Computer Vision Workshops}, pages 81--89, 2020{\natexlab{b}}.

\bibitem[Albiero et~al.(2020{\natexlab{c}})Albiero, Zhang, and
  Bowyer]{albiero2020does-balance_training}
V{\'\i}tor Albiero, Kai Zhang, and Kevin~W Bowyer.
\newblock How does gender balance in training data affect face recognition
  accuracy?
\newblock In \emph{Proceedings of the IEEE International Joint Conference on
  Biometrics}, pages 1--10. IEEE, 2020{\natexlab{c}}.

\bibitem[Albiero et~al.(2021)Albiero, Chen, Yin, Pang, and
  Hassner]{albiero2021img2pose}
Vitor Albiero, Xingyu Chen, Xi~Yin, Guan Pang, and Tal Hassner.
\newblock img2pose: Face alignment and detection via 6dof, face pose
  estimation.
\newblock In \emph{Proceedings of the IEEE/CVF Conference on Computer Vision
  and Pattern Recognition}, pages 7617--7627, 2021.

\bibitem[Albiero et~al.(2022)Albiero, Zhang, King, and Bowyer]{9650887}
Vítor Albiero, Kai Zhang, Michael~C. King, and Kevin~W. Bowyer.
\newblock Gendered differences in face recognition accuracy explained by
  hairstyles, makeup, and facial morphology.
\newblock 17:\penalty0 127--137, 2022.
\newblock \doi{10.1109/TIFS.2021.3135750}.

\bibitem[An et~al.(2021)An, Zhu, Gao, Xiao, Zhao, Feng, Wu, Qin, Zhang, Zhang,
  et~al.]{an2021partial}
Xiang An, Xuhan Zhu, Yuan Gao, Yang Xiao, Yongle Zhao, Ziyong Feng, Lan Wu, Bin
  Qin, Ming Zhang, Debing Zhang, et~al.
\newblock Partial fc: Training 10 million identities on a single machine.
\newblock In \emph{Proceedings of the IEEE/CVF International Conference on
  Computer Vision}, pages 1445--1449, 2021.

\bibitem[Bhatta et~al.(2023)Bhatta, Albiero, Bowyer, and
  King]{bhatta2023gender}
Aman Bhatta, V{\'\i}tor Albiero, Kevin~W Bowyer, and Michael~C King.
\newblock The gender gap in face recognition accuracy is a hairy problem.
\newblock In \emph{Proceedings of the IEEE/CVF Winter Conference on
  Applications of Computer Vision}, pages 303--312, 2023.

\bibitem[Cao et~al.(2018)Cao, Shen, Xie, Parkhi, and Zisserman]{vggface2}
Qiong Cao, Li~Shen, Weidi Xie, Omkar~M Parkhi, and Andrew Zisserman.
\newblock Vggface2: A dataset for recognising faces across pose and age.
\newblock In \emph{proceeding of the IEEE international Conference on Automatic
  Face and Gesture Recognition}, pages 67--74. IEEE, 2018.

\bibitem[Deng et~al.(2019)Deng, Guo, Xue, and Zafeiriou]{arcface}
Jiankang Deng, Jia Guo, Niannan Xue, and Stefanos Zafeiriou.
\newblock Arcface: Additive angular margin loss for deep face recognition.
\newblock In \emph{Proceedings of the IEEE/CVF Conference on Computer Vision
  and Pattern Recognition}, pages 4690--4699, 2019.

\bibitem[Deng et~al.(2020)Deng, Guo, Liu, Gong, and Zafeiriou]{deng2020sub}
Jiankang Deng, Jia Guo, Tongliang Liu, Mingming Gong, and Stefanos Zafeiriou.
\newblock Sub-center arcface: Boosting face recognition by large-scale noisy
  web faces.
\newblock In \emph{Proceedings of the European Conference on Computer Vision},
  pages 741--757. Springer, 2020.

\bibitem[Doctorow(2019)]{NIST_2019}
C.~Doctorow.
\newblock {NIST} confirms that facial recognition is a racist, sexist
  dumpster-fire, 2019.
\newblock https://boingboing.net/2019/12/19/demographics-v-robots.html5.

\bibitem[Drozdowski et~al.(2020)Drozdowski, Rathgeb, Dantcheva, Damer, and
  Busch]{drozdowski2020demographic}
Pawel Drozdowski, Christian Rathgeb, Antitza Dantcheva, Naser Damer, and
  Christoph Busch.
\newblock Demographic bias in biometrics: A survey on an emerging challenge.
\newblock \emph{IEEE Transactions on Technology and Society}, 1\penalty0
  (2):\penalty0 89--103, 2020.

\bibitem[Drozdowski et~al.(2021)Drozdowski, Rathgeb, and
  Busch]{drozdowski2021watchlist}
Pawel Drozdowski, Christian Rathgeb, and Christoph Busch.
\newblock The watchlist imbalance effect in biometric face identification:
  Comparing theoretical estimates and empiric measurements.
\newblock In \emph{Proceedings of the IEEE/CVF International Conference on
  Computer Vision}, pages 3757--3765, 2021.

\bibitem[Gross et~al.(2010)Gross, Matthews, Cohn, Kanade, and
  Baker]{gross2010multi}
Ralph Gross, Iain Matthews, Jeffrey Cohn, Takeo Kanade, and Simon Baker.
\newblock Multi-pie.
\newblock \emph{Image and vision computing}, 28\penalty0 (5):\penalty0
  807--813, 2010.

\bibitem[Grother()]{FRVT_2022}
Patrick Grother.
\newblock {NISTIR 8429}: Face recognition vendor test (frvt) part 8:
  Summarizing demographic differentials.
\newblock Technical report.

\bibitem[Grother(2022)]{NIST_2022}
Patrick Grother.
\newblock Face recognition vendor test (frvt) part 8: Summarizing demographic
  differentials.
\newblock 2022.

\bibitem[Grother and Phillips(2004)]{grother2004models}
Patrick Grother and P~Jonathon Phillips.
\newblock Models of large population recognition performance.
\newblock In \emph{Proceedings of the IEEE/CVF Conference on Computer Vision
  and Pattern Recognition}, volume~2, pages II--II. IEEE, 2004.

\bibitem[Guo()]{insightface}
Jia Guo.
\newblock Insightface: {2D} and {3D} face analysis project.
\newblock https://github.com/deepinsight/insightface, last accessed on February
  2021.

\bibitem[Guo et~al.(2016)Guo, Zhang, Hu, He, and Gao]{guo2016ms}
Yandong Guo, Lei Zhang, Yuxiao Hu, Xiaodong He, and Jianfeng Gao.
\newblock Ms-celeb-1m: A dataset and benchmark for large-scale face
  recognition.
\newblock In \emph{Proceedings of the European Conference on Computer Vision},
  pages 87--102. Springer, 2016.

\bibitem[Gwilliam et~al.(2021)Gwilliam, Hegde, Tinubu, and
  Hanson]{gwilliam2021rethinking}
Matthew Gwilliam, Srinidhi Hegde, Lade Tinubu, and Alex Hanson.
\newblock Rethinking common assumptions to mitigate racial bias in face
  recognition datasets.
\newblock In \emph{Proceedings of the IEEE/CVF International Conference on
  Computer Vision}, pages 4123--4132, 2021.

\bibitem[Han et~al.(2013)Han, Shan, Chen, and Gao]{han2013comparative}
Hu~Han, Shiguang Shan, Xilin Chen, and Wen Gao.
\newblock A comparative study on illumination preprocessing in face
  recognition.
\newblock \emph{Pattern Recognition}, 46\penalty0 (6):\penalty0 1691--1699,
  2013.

\bibitem[Hazirbas et~al.(2021)Hazirbas, Bitton, Dolhansky, Pan, Gordo, and
  Ferrer]{hazirbas2021towards}
Caner Hazirbas, Joanna Bitton, Brian Dolhansky, Jacqueline Pan, Albert Gordo,
  and Cristian~Canton Ferrer.
\newblock Towards measuring fairness in ai: the casual conversations dataset.
\newblock \emph{IEEE Transactions on Biometrics, Behavior, and Identity
  Science}, 4\penalty0 (3):\penalty0 324--332, 2021.

\bibitem[Hernandez-Ortega et~al.(2020)Hernandez-Ortega, Galbally, Fi{\'e}rrez,
  and Beslay]{faceqnetv1}
Javier Hernandez-Ortega, Javier Galbally, Julian Fi{\'e}rrez, and Laurent
  Beslay.
\newblock Biometric quality: Review and application to face recognition with
  faceqnet.
\newblock \emph{arXiv preprint arXiv:2006.03298}, 2020.

\bibitem[Hoggins(2019)]{Hoggins2019}
T.~Hoggins.
\newblock ‘racist and sexist’ facial recognition cameras could lead to
  false arrests, Dec. 20 2019.
\newblock
  https://www.telegraph.co.uk/technology/2019/12/20/racist-sexist-facial-recognition-cameras-could-lead-false-arrests/.

\bibitem[Huang et~al.(2008)Huang, Mattar, Berg, and Learned-Miller]{lfw}
Gary~B Huang, Marwan Mattar, Tamara Berg, and Eric Learned-Miller.
\newblock Labeled faces in the wild: A database forstudying face recognition in
  unconstrained environments.
\newblock In \emph{Workshop on faces in'Real-Life'Images: detection, alignment,
  and recognition}, 2008.

\bibitem[Hupont and Fern{\'a}ndez(2019)]{demogpairs}
Isabelle Hupont and Carles Fern{\'a}ndez.
\newblock Demogpairs: Quantifying the impact of demographic imbalance in deep
  face recognition.
\newblock In \emph{Proceedings of the IEEE International Conference on
  Automatic Face and Gesture Recognition}, pages 1--7. IEEE, 2019.

\bibitem[K{\"a}rkk{\"a}inen and Joo(2019)]{karkkainen2019fairface}
Kimmo K{\"a}rkk{\"a}inen and Jungseock Joo.
\newblock Fairface: Face attribute dataset for balanced race, gender, and age.
\newblock \emph{arXiv preprint arXiv:1908.04913}, 2019.

\bibitem[Kemelmacher-Shlizerman et~al.(2016)Kemelmacher-Shlizerman, Seitz,
  Miller, and Brossard]{megaface}
Ira Kemelmacher-Shlizerman, Steven~M Seitz, Daniel Miller, and Evan Brossard.
\newblock The megaface benchmark: 1 million faces for recognition at scale.
\newblock In \emph{Proceedings of the IEEE Conference on Computer Vision and
  Pattern Recognition}, pages 4873--4882, 2016.

\bibitem[Klare et~al.(2012)Klare, Burge, Klontz, Bruegge, and
  Jain]{klare2012face-age}
Brendan~F Klare, Mark~J Burge, Joshua~C Klontz, Richard W~Vorder Bruegge, and
  Anil~K Jain.
\newblock Face recognition performance: Role of demographic information.
\newblock \emph{IEEE Transactions on Information Forensics and Security},
  7\penalty0 (6):\penalty0 1789--1801, 2012.

\bibitem[Krishnapriya et~al.(2020)Krishnapriya, Albiero, Vangara, King, and
  Bowyer]{krishnapriya2020issues}
KS~Krishnapriya, V{\'\i}tor Albiero, Kushal Vangara, Michael~C King, and
  Kevin~W Bowyer.
\newblock Issues related to face recognition accuracy varying based on race and
  skin tone.
\newblock \emph{IEEE Transactions on Technology and Society}, 1\penalty0
  (1):\penalty0 8--20, 2020.

\bibitem[Krishnendu()]{krishnenduanalysis}
KS~Krishnendu.
\newblock Analysis of recent trends in face recognition systems.

\bibitem[Lema{\^\i}tre et~al.(2017)Lema{\^\i}tre, Nogueira, and
  Aridas]{lemaitre2017imbalanced}
Guillaume Lema{\^\i}tre, Fernando Nogueira, and Christos~K Aridas.
\newblock Imbalanced-learn: A python toolbox to tackle the curse of imbalanced
  datasets in machine learning.
\newblock \emph{The Journal of Machine Learning Research}, 18\penalty0
  (1):\penalty0 559--563, 2017.

\bibitem[Lohr(2018)]{Lohr2018}
S.~Lohr.
\newblock Facial recognition is accurate, if you’re a white guy.
\newblock \emph{The New York Times}, Feb. 9 2018.

\bibitem[Maze et~al.(2018)Maze, Adams, Duncan, Kalka, Miller, Otto, Jain,
  Niggel, Anderson, Cheney, et~al.]{ijbc}
Brianna Maze, Jocelyn Adams, James~A Duncan, Nathan Kalka, Tim Miller, Charles
  Otto, Anil~K Jain, W~Tyler Niggel, Janet Anderson, Jordan Cheney, et~al.
\newblock Iarpa janus benchmark-c: Face dataset and protocol.
\newblock In \emph{Proceedings of the international conference on biometrics},
  pages 158--165. IEEE, 2018.

\bibitem[Meng et~al.(2021)Meng, Zhao, Huang, and Zhou]{magface}
Qiang Meng, Shichao Zhao, Zhida Huang, and Feng Zhou.
\newblock {MagFace}: A universal representation for face recognition and
  quality assessment.
\newblock In \emph{Proceedings of the IEEE/CVF Conference on Computer Vision
  and Pattern Recognition}, 2021.

\bibitem[Moschoglou et~al.(2017)Moschoglou, Papaioannou, Sagonas, Deng, Kotsia,
  and Zafeiriou]{agedb}
Stylianos Moschoglou, Athanasios Papaioannou, Christos Sagonas, Jiankang Deng,
  Irene Kotsia, and Stefanos Zafeiriou.
\newblock Agedb: the first manually collected, in-the-wild age database.
\newblock In \emph{Proceedings of the IEEE Conference on Computer Vision and
  Pattern Recognition workshops}, pages 51--59, 2017.

\bibitem[Narayan et~al.(2021)Narayan, Molino, Goel, Neiswanger, and
  Re]{narayan2021personalized}
Avanika Narayan, Piero Molino, Karan Goel, Willie Neiswanger, and Christopher
  Re.
\newblock Personalized benchmarking with the ludwig benchmarking toolkit.
\newblock \emph{arXiv preprint arXiv:2111.04260}, 2021.

\bibitem[Parkhi et~al.(2015)Parkhi, Vedaldi, and Zisserman]{parkhi2015deep}
Omkar~M Parkhi, Andrea Vedaldi, and Andrew Zisserman.
\newblock Deep face recognition.
\newblock 2015.

\bibitem[Porgali et~al.(2023)Porgali, Albiero, Ryda, Ferrer, and
  Hazirbas]{porgali2023casual}
Bilal Porgali, V{\'\i}tor Albiero, Jordan Ryda, Cristian~Canton Ferrer, and
  Caner Hazirbas.
\newblock The casual conversations v2 dataset.
\newblock \emph{arXiv preprint arXiv:2303.04838}, 2023.

\bibitem[Ram et~al.(2010)Ram, Jalal, Jalal, and Kumar]{dbscan}
Anant Ram, Sunita Jalal, Anand~S Jalal, and Manoj Kumar.
\newblock A density based algorithm for discovering density varied clusters in
  large spatial databases.
\newblock \emph{International Journal of Computer Applications}, 3\penalty0
  (6):\penalty0 1--4, 2010.

\bibitem[Ricanek and Tesafaye(2006)]{ricanek2006morph}
Karl Ricanek and Tamirat Tesafaye.
\newblock Morph: A longitudinal image database of normal adult age-progression.
\newblock In \emph{Proceedings of the International Conference on Automatic
  Face and Gesture Recognition}, pages 341--345. IEEE, 2006.

\bibitem[Robbins et~al.(2023)Robbins, Zhou, Bhatta, Mello, Albiero, Bowyer, and
  Boult]{robbins2023cast}
Wes Robbins, Steven Zhou, Aman Bhatta, Chad Mello, V{\'\i}tor Albiero, Kevin~W
  Bowyer, and Terrance~E Boult.
\newblock Cast: Conditional attribute subsampling toolkit for fine-grained
  evaluation.
\newblock In \emph{Proceedings of the IEEE/CVF Winter Conference on
  Applications of Computer Vision}, pages 919--929, 2023.

\bibitem[Robinson et~al.(2020)Robinson, Livitz, Henon, Qin, Fu, and
  Timoner]{bfw}
Joseph~P Robinson, Gennady Livitz, Yann Henon, Can Qin, Yun Fu, and Samson
  Timoner.
\newblock Face recognition: too bias, or not too bias?
\newblock In \emph{Proceedings of the IEEE/CVF Conference on Computer Vision
  and Pattern Recognition Workshops}, pages 0--1, 2020.

\bibitem[Sengupta et~al.(2016)Sengupta, Chen, Castillo, Patel, Chellappa, and
  Jacobs]{cfp}
Soumyadip Sengupta, Jun-Cheng Chen, Carlos Castillo, Vishal~M Patel, Rama
  Chellappa, and David~W Jacobs.
\newblock Frontal to profile face verification in the wild.
\newblock In \emph{Proceedings of the IEEE winter conference on applications of
  computer vision}, pages 1--9. IEEE, 2016.

\bibitem[Sim et~al.(2003)Sim, Baker, and Bsat]{Sim2003PAMI}
Terence Sim, Simon Baker, and Maan Bsat.
\newblock The cmu pose, illumination, and expression database.
\newblock \emph{IEEE Transactions on Pattern Analysis and Machine
  Intelligence}, 25\penalty0 (12):\penalty0 1615--1618, 2003.

\bibitem[Terh{\"o}rst et~al.(2023)Terh{\"o}rst, Ihlefeld, Huber, Damer,
  Kirchbuchner, Raja, and Kuijper]{terhorst2023qmagface}
Philipp Terh{\"o}rst, Malte Ihlefeld, Marco Huber, Naser Damer, Florian
  Kirchbuchner, Kiran Raja, and Arjan Kuijper.
\newblock Qmagface: Simple and accurate quality-aware face recognition.
\newblock In \emph{Proceedings of the IEEE/CVF Winter Conference on
  Applications of Computer Vision}, pages 3484--3494, 2023.

\bibitem[Vangara et~al.(2019)Vangara, King, Albiero, Bowyer,
  et~al.]{vangara2019characterizing}
Kushal Vangara, Michael~C King, Vitor Albiero, Kevin Bowyer, et~al.
\newblock Characterizing the variability in face recognition accuracy relative
  to race.
\newblock In \emph{Proceedings of the IEEE/CVF Conference on Computer Vision
  and Pattern Recognition Workshops}, pages 0--0, 2019.

\bibitem[Wang et~al.(2021{\natexlab{a}})Wang, Liu, Hu, Shi, and
  Mei]{wang2021facex}
Jun Wang, Yinglu Liu, Yibo Hu, Hailin Shi, and Tao Mei.
\newblock Facex-zoo: A pytorch toolbox for face recognition.
\newblock In \emph{Proceedings of the 29th ACM International Conference on
  Multimedia}, pages 3779--3782, 2021{\natexlab{a}}.

\bibitem[Wang et~al.(2019)Wang, Deng, Hu, Tao, and Huang]{Wang_2019_ICCV}
Mei Wang, Weihong Deng, Jiani Hu, Xunqiang Tao, and Yaohai Huang.
\newblock Racial faces in the wild: Reducing racial bias by information
  maximization adaptation network.
\newblock In \emph{Proceedings of the International Conference on Computer
  Vision}, October 2019.

\bibitem[Wang et~al.(2021{\natexlab{b}})Wang, Zhang, and Deng]{wang2021meta}
Mei Wang, Yaobin Zhang, and Weihong Deng.
\newblock Meta balanced network for fair face recognition.
\newblock \emph{IEEE Transactions on Pattern Analysis and Machine
  Intelligence}, 2021{\natexlab{b}}.

\bibitem[Wang et~al.(2021{\natexlab{c}})Wang, Zhang, Xiong, and
  Zhao]{wang2021face}
Qingzhong Wang, Pengfei Zhang, Haoyi Xiong, and Jian Zhao.
\newblock Face. evolve: A high-performance face recognition library.
\newblock \emph{arXiv preprint arXiv:2107.08621}, 2021{\natexlab{c}}.

\bibitem[Wu et~al.(2023{\natexlab{a}})Wu, Albiero, Krishnapriya, King, and
  Bowyer]{wu2023face}
Haiyu Wu, V{\'\i}tor Albiero, KS~Krishnapriya, Michael~C King, and Kevin~W
  Bowyer.
\newblock Face recognition accuracy across demographics: Shining a light into
  the problem.
\newblock In \emph{Proceedings of the IEEE/CVF Conference on Computer Vision
  and Pattern Recognition}, pages 1041--1050, 2023{\natexlab{a}}.

\bibitem[Wu et~al.(2023{\natexlab{b}})Wu, Bezold, Bhatta, and
  Bowyer]{wu2023logical}
Haiyu Wu, Grace Bezold, Aman Bhatta, and Kevin~W Bowyer.
\newblock Logical consistency and greater descriptive power for facial hair
  attribute learning.
\newblock In \emph{Proceedings of the IEEE/CVF Conference on Computer Vision
  and Pattern Recognition}, pages 8588--8597, 2023{\natexlab{b}}.

\bibitem[Yi et~al.(2014)Yi, Lei, Liao, and Li]{yi2014learning}
Dong Yi, Zhen Lei, Shengcai Liao, and Stan~Z Li.
\newblock Learning face representation from scratch.
\newblock \emph{arXiv preprint arXiv:1411.7923}, 2014.

\bibitem[Yu et~al.(2018)Yu, Wang, Peng, Gao, Yu, and Sang]{yu2018bisenet}
Changqian Yu, Jingbo Wang, Chao Peng, Changxin Gao, Gang Yu, and Nong Sang.
\newblock Bisenet: Bilateral segmentation network for real-time semantic
  segmentation.
\newblock In \emph{Proceedings of the European Conference on Computer Vision},
  pages 325--341, 2018.

\bibitem[Zheng and Deng(2018)]{cplfw}
Tianyue Zheng and Weihong Deng.
\newblock Cross-pose lfw: A database for studying cross-pose face recognition
  in unconstrained environments.
\newblock \emph{Beijing University of Posts and Telecommunications, Tech. Rep},
  5\penalty0 (7), 2018.

\bibitem[Zheng et~al.(2017)Zheng, Deng, and Hu]{calfw}
Tianyue Zheng, Weihong Deng, and Jiani Hu.
\newblock Cross-age lfw: A database for studying cross-age face recognition in
  unconstrained environments.
\newblock \emph{arXiv preprint arXiv:1708.08197}, 2017.

\bibitem[Zhu and Mart{\'\i}nez(2004)]{zhu2004optimal}
Manli Zhu and Aleix~M Mart{\'\i}nez.
\newblock Optimal subclass discovery for discriminant analysis.
\newblock In \emph{Proceedings of the IEEE/CVF Conference on Computer Vision
  and Pattern Recognition Workshop}, pages 97--97. IEEE, 2004.

\bibitem[Zhu and Martinez(2006)]{zhu2006subclass}
Manli Zhu and Aleix~M Martinez.
\newblock Subclass discriminant analysis.
\newblock \emph{IEEE Transactions on Pattern Analysis and Machine
  Intelligence}, 28\penalty0 (8):\penalty0 1274--1286, 2006.

\end{thebibliography}
\end{document}